\documentclass{article}
\usepackage{iclr2027_conference,times}

\usepackage{amsmath,amssymb,amsthm}
\usepackage{booktabs}
\usepackage{array}
\usepackage{graphicx}
\usepackage{float}
\usepackage{xcolor}
\usepackage{url}
\usepackage[capitalise,noabbrev]{cleveref}

\newtheorem{proposition}{Proposition}
\newcommand{\bobname}{BoB}
\newcommand{\bobsupport}{BoB-Support}
\newcommand{\bobconstant}{BoB-Constant}
\newcommand{\corpusname}{WildScores}

\newcommand{\bobMedianPairDistTwo}{0.89}

\newcommand{\bobTauPairDistThree}{0.509}
\newcommand{\bobTauPairStrippedDistThree}{0.647}

\newcommand{\bobOutIndexLevelThree}{0.939}
\newcommand{\bobOutIndexProfileThree}{0.632}
\newcommand{\bobAggregationProfileThree}{0.483}
\newcommand{\bobAggregationLevelThree}{0.904}

\newcommand{\bobNeighborMassMinThree}{0.004}
\newcommand{\bobNeighborMassMaxThree}{0.180}
\newcommand{\bobAimeNeighborMassMinThree}{0.482}
\newcommand{\bobAimeNeighborMassMaxThree}{0.532}
\newcommand{\bobDensityMassAtMedianOne}{1.5}

\newcommand{\bobHeadlineProfileThree}{0.483}

\newcommand{\bobFlatProfileThree}{0.473}
\newcommand{\bobFlatProfilePThree}{0.121}
\newcommand{\bobFlatCopyShiftThree}{0.963}
\newcommand{\bobCopyShiftThree}{0.095}
\newcommand{\bobShuffleMeanThree}{0.358}

\newcommand{\bobShuffleMaxThree}{0.447}
\newcommand{\bobLatentLevelThree}{0.907}
\newcommand{\bobLatentProfileThree}{0.459}
\newcommand{\bobInfoLevelThree}{0.907}
\newcommand{\bobInfoProfileThree}{0.484}
\newcommand{\bobInfoProfilePTwo}{1.00}
\newcommand{\bobParaLevelThree}{0.905}
\newcommand{\bobParaProfileThree}{0.493}
\newcommand{\bobStrippedLevelThree}{0.903}
\newcommand{\bobStrippedProfileThree}{0.475}

\newcommand{\bobThetaIsotonicThree}{0.381}
\newcommand{\bobThetaIsotonicPFour}{0.0026}
\newcommand{\bobStrictIsotonicThree}{0.464}
\newcommand{\bobOutIndexAAProfileThree}{0.371}
\newcommand{\bobNonSingletonProfileThree}{0.085}

\newcommand{\bobNonSingletonProfilePThree}{0.076}

\newcommand{\bobCopyFixedRankOne}{10.4}
\newcommand{\bobCopyFullRankOne}{15.9}
\newcommand{\bobAdjacentGapFour}{0.0042}
\newcommand{\bobRobustLevelBoundThree}{0.004}
\newcommand{\bobRobustProfileBoundThree}{0.007}

\newcommand{\bobEmbeddingSmallCopyRankOne}{24.1}
\newcommand{\bobEmbeddingMediumCopyRankOne}{15.8}
\newcommand{\bobEmbeddingLargeCopyRankOne}{15.9}

\newcommand{\bobAASnapshotRows}{643}
\newcommand{\bobAARetainedRows}{605}
\newcommand{\bobPopulationMinAdded}{43}
\newcommand{\bobPopulationMaxAdded}{81}
\newcommand{\bobClosePairHolmSurvivors}{equal weighting, nearest benchmark, and first principal component}
\newcommand{\bobEmbeddingHeadlineSpreadBoundThree}{0.002}
\newcommand{\bobAASnapshotRetrieved}{2026-09-03T20:19:24Z}
\newcommand{\bobAASnapshotSha}{f2a230e9bd1fd55bfe058eaa59ae3f0401dbde76934609f4c5eef0adf57f745a}

\newcommand{\bobReleaseBenchmarks}{148}

\newcommand{\bobReleaseReviewedConfigs}{354}
\newcommand{\bobReleaseReviewedGroups}{316}
\newcommand{\bobReleaseReviewedCells}{5,371}
\newcommand{\bobReleaseReviewedObservedPercent}{10.3}

\newcommand{\bobReleaseBobRegretThree}{1.67}

\newcommand{\bobReleaseThetaRank}{0.817}

\newcommand{\bobReleaseThetaRegretThree}{1.62}

\newcommand{\bobReleaseUniformRegretThree}{2.08}

\newcommand{\bobReleaseReviewedThetaRankDelta}{+0.006}
\newcommand{\bobReleaseReviewedThetaRankLo}{+0.003}
\newcommand{\bobReleaseReviewedThetaRankHi}{+0.011}

\newcommand{\bobReleaseReviewedThetaRegretDelta}{+0.055}

\newcommand{\bobReleaseReviewedThetaRegretHolm}{1.000}

\newcommand{\bobReleaseReviewedMatchedRankHolm}{0.486}

\newcommand{\bobReleaseReviewedConstantRankHolm}{0.006}

\newcommand{\bobReleaseRankTargets}{133}
\newcommand{\bobReleaseRankFamilies}{67}

\newcommand{\bobAAIndexProfileHolm}{0.292}

\newcommand{\bobMatchedCuratedRank}{0.742}
\newcommand{\bobMatchedPCRank}{0.631}
\newcommand{\bobMatchedCovarianceRank}{0.423}
\newcommand{\bobMatchedThetaRank}{0.802}
\newcommand{\bobMatchedSupportRank}{0.811}

\newcommand{\bobMatchedTargets}{120}
\newcommand{\bobMatchedFamilies}{60}

\newcommand{\bobMatchedThetaCuratedDelta}{+0.060}
\newcommand{\bobMatchedThetaCuratedLo}{+0.038}
\newcommand{\bobMatchedThetaCuratedHi}{+0.081}
\newcommand{\bobMatchedUniformCuratedDelta}{+0.008}
\newcommand{\bobMatchedUniformCuratedP}{0.779}

\newcommand{\bobStrictTargets}{50}
\newcommand{\bobStrictFamilies}{29}

\newcommand{\bobStrictRetention}{+0.0350}
\newcommand{\bobStrictRetentionHolm}{0.0130}
\newcommand{\bobStrictRetentionLo}{0.0183}
\newcommand{\bobStrictRetentionHi}{0.0520}
\newcommand{\bobStrictFlatRetention}{+0.0367}

\newcommand{\bobStrictDensityDelta}{-0.0016}

\newcommand{\bobStrictOutsideRetention}{+0.0373}
\newcommand{\bobStrictOutsideHolm}{0.0673}
\newcommand{\bobLowTargets}{120}
\newcommand{\bobLowFamilies}{60}

\newcommand{\bobLowFullPC}{0.652}

\newcommand{\bobLowFullMasked}{0.779}

\newcommand{\bobLowFullAffine}{0.815}

\newcommand{\bobLowRetention}{+0.0364}
\newcommand{\bobLowRetentionHolm}{0.0107}

\newcommand{\bobLowFlatRetention}{+0.0342}

\newcommand{\bobLowDensityDelta}{+0.0022}

\newcommand{\bobLowOutsideRetention}{+0.0379}
\newcommand{\bobLowOutsideHolm}{0.0136}
\newcommand{\bobListingUniform}{5.907}
\newcommand{\bobListingFlat}{1.948}
\newcommand{\bobListingDensity}{0.626}
\newcommand{\bobListingIdentity}{0.150}

\newcommand{\bobAbstractListingFlat}{1.95}
\newcommand{\bobAbstractListingDensity}{0.63}
\newcommand{\bobListingReduction}{67.9}
\newcommand{\bobListingParaphrase}{0.955}

\newcommand{\bobListingIdentityMaxChange}{3.4\times10^{-8}}
\newcommand{\bobListingExactMax}{44}
\newcommand{\bobListingParaphraseMax}{55}

\newcommand{\bobMetadataExcluded}{38}
\newcommand{\bobMetadataRetained}{110}
\newcommand{\bobMetadataExcludedCells}{1,750}
\newcommand{\bobMetadataRetainedCells}{3,621}

\newcommand{\bobMetadataStrictTargets}{3}
\newcommand{\bobMetadataStrictFamilies}{3}

\newcommand{\bobMetadataLowTargets}{66}
\newcommand{\bobMetadataLowFamilies}{43}
\newcommand{\bobMetadataLowRetention}{+0.0241}
\newcommand{\bobMetadataLowHolm}{0.6093}
\newcommand{\bobMetadataLowRetentionLo}{0.002}
\newcommand{\bobMetadataLowRetentionHi}{0.050}

\newcommand{\bobPartialFullDensity}{0.625}

\newcommand{\bobPartialQuarterFlat}{0.684}
\newcommand{\bobPartialQuarterDensity}{0.761}
\newcommand{\bobPartialQuarterUnexposedWithin}{0.668}
\newcommand{\bobPartialQuarterUnexposedFit}{0.096}

\newcommand{\bobSupportedFullAffine}{1.041}
\newcommand{\bobSupportedFullDensityAffine}{0.220}
\newcommand{\bobSupportedQuarterAffine}{0.546}
\newcommand{\bobSupportedQuarterDensityAffine}{0.567}

\newcommand{\bobSemanticMass}{104.43}
\newcommand{\bobSemanticGroups}{104}
\newcommand{\bobSemanticHalfGroups}{141}
\newcommand{\bobOverlapNearestMedian}{36.1}
\newcommand{\bobOverlapNearestLow}{66}

\newcommand{\bobOverlapStrongPairs}{8}
\newcommand{\bobOverlapStrongBenchmarks}{13}

\newcommand{\bobFamilyPositive}{28}
\newcommand{\bobFamilyNegative}{15}
\newcommand{\bobFamilyMedian}{+0.0173}

\newcommand{\bobFamilyDeleteLo}{+0.0152}
\newcommand{\bobFamilyDeleteHi}{+0.0281}

\newcommand{\bobSemanticStrictDelta}{+0.0093}
\newcommand{\bobSemanticStrictLo}{+0.0011}
\newcommand{\bobSemanticStrictHi}{+0.0183}
\newcommand{\bobSemanticLowDelta}{+0.0036}
\newcommand{\bobSemanticLowLo}{-0.0004}
\newcommand{\bobSemanticLowHi}{+0.0080}
\newcommand{\bobSemanticExactDensity}{0.625}
\newcommand{\bobSemanticExactGroup}{0.876}
\newcommand{\bobSemanticParaphraseDensity}{0.953}
\newcommand{\bobSemanticParaphraseGroup}{1.440}
\newcommand{\bobMatchedOriginalCurated}{0.7433}
\newcommand{\bobMatchedOriginalFull}{0.7941}
\newcommand{\bobMatchedOriginalGap}{+0.0508}
\newcommand{\bobMatchedRestrictedCurated}{0.7908}
\newcommand{\bobMatchedRestrictedFull}{0.8149}
\newcommand{\bobMatchedRestrictedGap}{+0.0241}
\newcommand{\bobMatchedProtocolTargets}{66}

\newcommand{\bobRulerMinCoefficient}{0.388}
\newcommand{\bobRulerMaxCoefficient}{0.497}
\newcommand{\bobRulerGroupCoefficient}{2.581}

\newcommand{\bobSparseDensity}{0.0843}
\newcommand{\bobSparseEqual}{0.0852}
\newcommand{\bobSparseDenseReference}{0.2298}
\newcommand{\bobSparseGap}{-0.0009}
\newcommand{\bobSparseGapLo}{-0.0041}
\newcommand{\bobSparseGapHi}{0.0022}
\newcommand{\bobRulerRankAgreement}{0.9971}
\newcommand{\bobRulerHeldoutDensity}{0.2180}
\newcommand{\bobRulerHeldoutEqual}{0.2215}

\newcommand{\bobRulerRelatedDensity}{0.0047}
\newcommand{\bobRulerRelatedEqual}{0.0198}

\newcommand{\bobMaskCopyGap}{-0.5332}
\newcommand{\bobMaskCopyLo}{-0.7286}
\newcommand{\bobMaskCopyHi}{-0.3508}
\newcommand{\bobMaskCopyHolm}{0.0117}
\newcommand{\bobMaskAffineGap}{-0.4006}
\newcommand{\bobMaskAffineLo}{-0.6122}
\newcommand{\bobMaskAffineHi}{-0.1900}
\newcommand{\bobMaskAffineHolm}{0.0156}
\newcommand{\bobMaskPredictionGap}{0.0000}
\newcommand{\bobMaskPredictionLo}{-0.0013}
\newcommand{\bobMaskPredictionHi}{0.0012}

\newcommand{\bobMaskRetentionGap}{-0.0008}
\newcommand{\bobMaskRetentionLo}{-0.0045}
\newcommand{\bobMaskRetentionHi}{0.0029}

\newcommand{\bobNatShareEqualBase}{33.8}
\newcommand{\bobNatShareEqualRemeasure}{44.1}
\newcommand{\bobNatShareModelBase}{33.7}
\newcommand{\bobNatShareModelShared}{35.9}
\newcommand{\bobNatShareModelSeparate}{39.4}
\newcommand{\bobNatDispEqual}{5.61}
\newcommand{\bobNatDispModelShared}{4.20}
\newcommand{\bobNatDispModelSeparate}{4.95}
\newcommand{\bobNatStrictGapShared}{+0.0023}
\newcommand{\bobNatStrictGapSharedLo}{-0.0037}
\newcommand{\bobNatStrictGapSharedHi}{+0.0090}
\newcommand{\bobNatStrictGapSeparate}{-0.0038}
\newcommand{\bobNatStrictGapSeparateLo}{-0.0072}
\newcommand{\bobNatStrictGapSeparateHi}{-0.0005}
\newcommand{\bobNatBroadGapShared}{+0.0047}
\newcommand{\bobNatBroadGapSharedLo}{+0.0017}
\newcommand{\bobNatBroadGapSharedHi}{+0.0079}
\newcommand{\bobNatBroadGapSeparate}{+0.0030}
\newcommand{\bobNatBroadGapSeparateLo}{+0.0008}
\newcommand{\bobNatBroadGapSeparateHi}{+0.0055}
\newcommand{\bobNatShareFullLow}{47.5}
\newcommand{\bobNatShareFullHigh}{59.9}

\newcommand{\bobNatFamilies}{24}
\newcommand{\bobNatFamilyColumns}{79}
\newcommand{\bobNatFamiliesModelSmaller}{16}
\newcommand{\bobNatFamiliesModelLarger}{4}
\newcommand{\bobNatFamiliesModelTies}{4}

\newcommand{\bobNatRulerEqual}{2.20}
\newcommand{\bobNatRulerColumn}{0.75}
\newcommand{\bobNatRulerModel}{0.97}
\newcommand{\bobNatAimeEqual}{10.82}
\newcommand{\bobNatAimeColumn}{6.98}
\newcommand{\bobNatAimeModel}{8.69}

\newcommand{\bobExternalModels}{30}
\newcommand{\bobExternalTasks}{143}
\newcommand{\bobExternalLeaves}{596}
\newcommand{\bobExternalTargets}{131}
\newcommand{\bobExternalCoverage}{99.6}

\newcommand{\bobExternalRetentionHolm}{0.0003}

\newcommand{\bobExternalDensityDelta}{+0.0038}
\newcommand{\bobExternalDensityLo}{-0.0007}
\newcommand{\bobExternalDensityHi}{+0.0084}
\newcommand{\bobExternalResidualAssociation}{0.0136}
\newcommand{\bobExternalResidualLo}{-0.0275}
\newcommand{\bobExternalResidualHi}{+0.0610}
\newcommand{\bobExternalReuseAP}{0.1462}
\newcommand{\bobExternalReusePrevalence}{0.0177}
\newcommand{\bobExternalRuleSelected}{192}
\newcommand{\bobExternalRuleTotal}{336}
\newcommand{\bobExternalRuleHarm}{15}
\newcommand{\bobExternalRuleRankDelta}{-0.1651}
\newcommand{\bobExternalRuleRankLo}{-0.2333}
\newcommand{\bobExternalRuleRankHi}{-0.0796}
\newcommand{\bobExternalRuleUtilityDelta}{+0.0084}
\newcommand{\bobExternalRuleUtilityLo}{-0.0021}
\newcommand{\bobExternalRuleUtilityHi}{+0.0195}
\newcommand{\bobExternalRuleUtilityLower}{-0.0005}

\title{Balance of Benchmarks:\\
Evidence Retention and\\
Multiplicity-Aware Model Comparison}

\author{JHEN-KE LIN\\
National Yang Ming Chiao Tung University\\
\texttt{jacob.cs14@nycu.edu.tw}
\And
Hong-Yun Lin\\
National Taiwan University\\
\texttt{r15922020@ntu.edu.tw}}

\ifdefined\bobAnonymousSubmission
\else
\iclrpreprintcopy
\fi

\begin{document}

\maketitle

\begin{abstract}
Benchmark aggregation requires decisions about how much evidence to use and
how much influence each benchmark receives. Curated suites control influence
through selection, which can exclude informative observations. A broader
pool can retain those observations. With equal weighting, repeated or
closely related benchmarks gain influence as their number grows.
We introduce Balance of Benchmarks (\bobname{}) to retain eligible
observations and control their influence separately. We fit
benchmark-specific response curves to place different difficulty ranges on
a shared ability scale and use description similarity to downweight
benchmarks in crowded neighborhoods. We test retention and density
weighting in separate interventions on \corpusname{}, a sparse collection
of \bobReleaseBenchmarks{} developer-reported benchmarks, using 14
Artificial Analysis benchmarks as a controlled reference. With the model
population fixed and entire source-lineage families held out, suite
expansion improves held-out prediction under \bobname{} and equal-weight
equating on both \corpusname{} and BIG-bench. Under substantial reporting
overlap, density weighting improves robustness to multiplicity. Complete
re-listings yield smaller mean rank displacement under \bobname{}
(\bobAbstractListingDensity{} versus \bobAbstractListingFlat{} for
equal-weight equating), with benefits also in paraphrased re-listings and
supported affine fits. The benefit can reverse if models report different
benchmarks, because a single weight per column gives models that report
only part of a group less than its intended share. We therefore compute
each model's weights from the benchmarks it reports. This rule removes the
reversal in controlled re-listings and keeps held-out prediction close to
that of equal weighting. These comparisons separate evidence retention,
benchmark influence, and reporting coverage and show why repeated entries
need to be handled per model.
\end{abstract}

\section{Introduction}\label{sec:intro}

Model comparison involves choosing which observations to use and how much
weight each benchmark should receive. Curated suites control
capability emphasis through benchmark selection and category weights, as in
the Artificial Analysis (AA) Intelligence Index
\citep{artificialanalysis2026methodology}. Selection can exclude informative
observations. A benchmark that saturates on leading models may still
distinguish models near the floor of a harder evaluation. We ask whether
aggregation can retain this additional evidence while controlling the
influence of repeated or closely related benchmarks.

We study benchmark-level scores already reported in public model releases.
These reports supply a sparse matrix whose columns differ in difficulty,
capability coverage, reporting frequency, and evaluation protocol. Benchmark
descriptions and provenance accompany the aggregate scores. At this
resolution, reporting coverage and score comparability shape the comparison.
To test prediction, we compare a global ordering with held-out reported
benchmark rankings among eligible release groups.

Listing a benchmark twice gives one set of measurements extra weight under
equal weighting. A related benchmark can also supply a fresh measurement,
which makes evidence collection and emphasis separate choices. Reporting
coverage then determines how much nominal
group weight a model receives. A fixed weight shared across related columns
gives different observed group weights to models evaluated on different
subsets. We therefore distinguish benchmark weights, model-specific
observed weights, and the resulting changes in fitted rankings.

Benchmark signatures map capability overlap using token-perplexity
patterns that predict benchmark scores \citep{wu2026overlaps}. To use such
relationships in model comparisons, we need a rule for assigning weights
and accounting for how those weights interact with missing observations.

In Balance of Benchmarks (\bobname{}), we admit scores with documented
provenance and a normalizable scale. We fit one response curve per
benchmark to place easy and hard benchmarks on a shared ability scale.
On that scale, inverse-density weighting assigns smaller coefficients to
benchmarks in crowded description neighborhoods. Benchmark-level response
models and clone-robust weighting have established precedents
\citep{ho2025rosetta,berriaud2026clone,berriaud2026consensus}. We study how
they interact under sparse reporting. For a fixed eligible list, benchmark
weights depend only on descriptions. Their effect on fitted ability
$\theta_m$ also depends on the fitted curves and which scores each model
reports. We describe and evaluate an optional task-conditioned extension in
Appendix~\ref{sec:app-task-extension}.

We test retention and influence control separately on \corpusname{}, with
AA as a denser reference. For retention, we expand a curated suite of ten
benchmarks from the historical AA Index v3.0 list
\citep{artificialanalysis2026methodology} to the full eligible pool. We
hold the model population fixed and withhold whole source-lineage families.
On these matched populations, held-out ranking improves under both
\bobname{} and equal-weight equating. For influence control, we re-list
existing benchmarks with the same complete scores and observation masks.
Density weighting reduces mean rank displacement by \bobListingReduction{}\%
relative to equal-weight equating, and the effect carries over to an
affine-factor estimator. We vary reporting and fitting conditions to test
when weighting yields stable comparisons. On a previously unused BIG-bench
score matrix, we test retention across suite policies and examine observable
overlap conditions for controlled re-listings. Its task lineage partly
overlaps with \corpusname{}.

\paragraph{Contributions.}
\textbf{(i) An integrated formulation for sparse reported scores.}
We specify evidence eligibility, difficulty equating, and semantic
weighting in one reproducible aggregation procedure.
\textbf{(ii) Separate tests of retention and influence control.}
Fixed-population held-out prediction and exact or paraphrased re-listings
distinguish evidence-pool gains from sensitivity to benchmark multiplicity.
We compare observed-cell factors and same-information semantic grouping
separately from the within-estimator retention contrast.
\textbf{(iii) Analysis of reporting effects and a reporting-aware weighting.}
Conserving a group's column weight can change each model's observed weight.
We derive this constraint for positive column-only rules and test its
consequences under two estimators. In controlled re-listings, evaluating
density over each model's reported benchmarks removes the observed
reversal. We provide the \corpusname{} scores and their associated source
records.

\section{Related work}\label{sec:related}

\paragraph{Benchmark selection and evidence efficiency.}
Holistic suites organize broad evaluations
\citep{liang2023holistic,srivastava2023bigbench}, and AA combines selected
benchmarks with prescribed weights \citep{artificialanalysis2026}.
Evaluation-compression methods select fewer measurements to preserve model
comparisons
\citep{perlitz2024efficient,polo2024tinybenchmarks,kipnis2025metabench,vivek2024anchor},
and marginal-coverage approaches use prompt representations or score
covariance to guide subset choice \citep{yao2026coresets,smola2026submodular}.
These methods select evidence for efficient evaluation. Once scores are
reported, we test evidence retention on fixed model populations and use
descriptions to assign benchmark weights.

\paragraph{Aggregation and benchmark influence.}
Arena aggregates human preferences \citep{chiang2024arena}, with
participation and selective disclosure affecting its evidence
\citep{singh2025leaderboardillusion}. Rank aggregation offers another way to
combine benchmark results \citep{colombo2022benchmarking}, domain-specific
taxonomies define explicit capability groups \citep{hudson2026metabench}, and
score-matrix methods study agreement, dimensionality, and predictability
\citep{perlitz2024agreement,ruan2024observational,benchscope2026}.
\citet{himmi2024missing} address missing scores through compatible partial
rankings and Borda aggregation. We study how reporting masks change the
influence assigned by description-based weights. Benchmark composition
also affects ranking sensitivity, with trade-offs involving benchmark
diversity \citep{zhang2024diversity}. Within benchmarks,
\citet{uzunoglu2026flaw} use benchmark Harmony to measure the distribution
of performance across subdomains. \citet{siska2024examining} show that
accounting for correlated test prompts can change rankings, with semantic
similarity among the explanatory factors. To control multiplicity,
\citet{berriaud2026clone,berriaud2026consensus} develop clone-robust task
weights from metric similarity, with axioms governing symmetry, continuity,
and locality. Game-theoretic evaluation uses tabulated payoffs to construct
ratings robust to task or prompt duplication
\citep{balduzzi2018reevaluating,liu2025reevaluating}.
We use a reciprocal kernel-density rule that shares weight across isolated
co-located entries (\Cref{sec:weighting}). Its effective mass is the
order-zero similarity-sensitive diversity of \citet{leinster2012diversity}
at uniform frequencies, with related sums in the spread construction of
\citet{willerton2015spread}. In sparse-score estimation, neighboring
descriptions, reporting masks, initialization, and refitting determine how
these column weights affect rankings.

\paragraph{From benchmark relationships to model comparison.}
\citet{wu2026overlaps} and \bobname{} both treat relationships among
benchmarks as part of evaluation breadth. Their signatures estimate
functional overlap from token-perplexity patterns predictive of benchmark
scores. These relationships matter for aggregation because related
benchmarks can supply additional measurements while warranting control of
their collective influence. We separate retention from influence
allocation, equate heterogeneous score scales, and analyze how missing
observations alter the weights each model receives. A signature-derived
kernel could supply the benchmark relation for this procedure. We
implement the benchmark-level instance using descriptions
(\Cref{sec:app-signatures,tab:signature-comparison}).

\paragraph{Score linking.}
To compare benchmarks with different difficulty and saturation, we use
characteristic curves to place scores on a common latent scale, following
score-equating ideas from item response theory
\citep{lalor2016irt,rodriguez2021evaluation,lord1968statistical,stocking1983metric,kolen2014equating}.
The Rosetta Stone framework underlying the Epoch Capabilities Index fits
model capability, benchmark difficulty, and discrimination to partially
observed aggregate scores, including externally reported evaluations
\citep{ho2025rosetta}. We combine this benchmark-level modeling approach with
description-based weighting. This combination supports fixed-population
retention comparisons and tests of how multiplicity affects rankings under
sparse reporting.

\section{Method}\label{sec:method}

\begin{figure}[t]
\centering
\includegraphics[width=\linewidth]{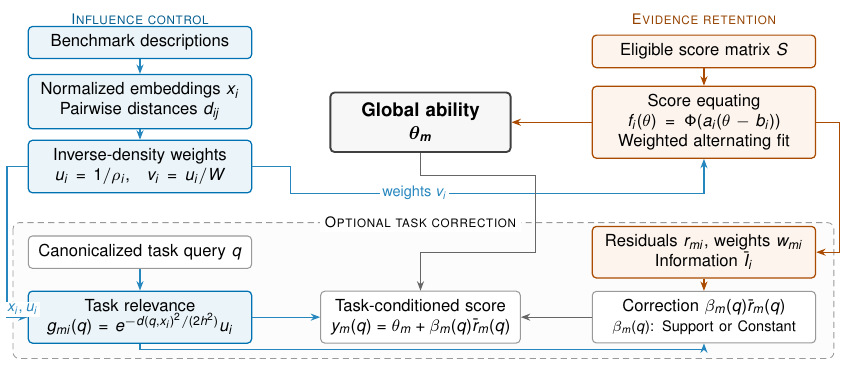}
\caption{\textbf{What evidence enters and how much it counts.}
Eligible scores enter score equating, while description-derived density
weights set benchmark coefficients in the fitting objective. Scores,
observation masks, curves, and robust cell weights jointly determine global
ability $\theta_m$. Held-out prediction tests retention, and re-listing
tests ranking sensitivity. The dashed branch is an optional extension
(Appendix~\ref{sec:app-task-extension}).}
\label{fig:pipeline}
\end{figure}

In \bobname{}, we combine score equating and semantic inverse-density
weighting on eligible scores $S=(s_{mi})$ (\Cref{fig:pipeline}).
Response curves link score scales, and description geometry sets benchmark
coefficients in the global fit. We construct the geometry by mapping
descriptions to unit vectors $x_i$ with Qwen3-Embedding-8B
\citep{qwen3embedding2025} and comparing them by angular distance
$d(\cdot,\cdot)$. This proximity represents related content. We assess
measurement dependence separately.

\paragraph{Eligible evidence.}
We include in $S$ scores with documented sources and evaluation settings,
normalized scales, and sufficient observations in their benchmark columns
and model rows (\Cref{sec:app-release-data}). To assess residual protocol
differences within a column, we apply a separate metadata restriction
(\Cref{sec:release-prediction}).

\subsection{Score equating}\label{sec:curves}

\begin{figure}[t]
\centering
\includegraphics[width=\linewidth]{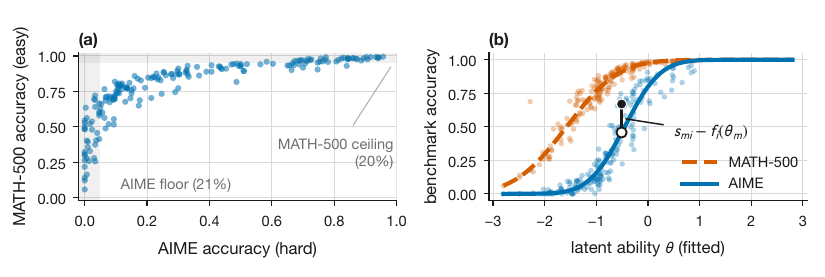}
\caption{\textbf{Benchmark-specific curves place different difficulty ranges
on a shared ability scale.}
\textbf{(a)} AIME's floor and MATH-500's ceiling in the frozen AA snapshot.
Shaded bands mark scores within 0.05 of the endpoints.
\textbf{(b)} Curves give expected scores at shared ability. Filled and open
markers show one model's observed and expected AIME scores, whose gap
divided by $\sigma_i$ gives $r_{mi}$ (\Cref{eq:icc}).}
\label{fig:saturation}
\end{figure}

To link the eligible scores, we place them on a shared latent ability
scale. AIME and MATH-500 illustrate how characteristic curves represent
saturation near zero and one (\Cref{fig:saturation}). We fit one monotone
expected-score curve per benchmark to aggregate scores, using a working
model inspired by item response theory
\citep{lord1968statistical,embretson2000irt}.
\begin{equation}\label{eq:icc}
 f_i(\theta)=\Phi\big(a_i(\theta-b_i)\big),\qquad
 r_{mi}=\frac{s_{mi}-f_i(\theta_m)}{\sigma_i}.
\end{equation}
Here $a_i$ and $b_i$ describe discrimination and difficulty, $\Phi$ is the
standard normal cumulative distribution function, and $\sigma_i$ is the
residual scale. A positive residual $r_{mi}$ means above-expected performance
at the model's fitted ability.

We alternate curve fitting with a weighted least-squares ability update
\begin{equation}\label{eq:theta}
 \theta_m=\arg\min_\theta\sum_{i:\,s_{mi}\ \mathrm{observed}}
 w_{mi}v_i\big(s_{mi}-f_i(\theta)\big)^2,
\end{equation}
where the robust weight $w_{mi}$ limits the influence of outlying scores and
$v_i$ is the semantic coefficient of \Cref{sec:weighting}.
After each ability update, we standardize ability and transform the curves
jointly as $\theta'=(\theta-\mu)/s$, $a'=sa$ and $b'=(b-\mu)/s$, preserving every
curve prediction. The fitted $\theta_m$ gives the model's global level.
Because monotone curves share one model ordering across benchmarks, we use
an optional residual correction to represent task-specific departures
(Appendix~\ref{sec:app-task-extension}).
\Cref{tab:fit-procedure} consolidates the full fitting procedure, with
score-weight definitions in \Cref{sec:app-score-weights}.
We stop after six rounds of curve fitting and ability updates. In the
fixed-population comparisons, full-pool held-out Spearman falls with 48
rounds (\Cref{tab:insight-longer}). \Cref{sec:app-release-stability} reports
additional stopping sensitivities.

\subsection{Semantic inverse-density weighting}\label{sec:weighting}

Equal weighting assigns a full coefficient to each listing of a
benchmark. To share weight across entries in crowded neighborhoods, we
weight each benchmark by the inverse of its local description density.
\begin{equation}\label{eq:mass}
 k(d)=e^{-d^2/(2h_{\mathrm{dens}}^2)},\qquad
 \rho_i=\sum_j k(d_{ij}),\qquad u_i=\rho_i^{-1},\qquad W=\sum_i u_i.
\end{equation}
An entry retains more mass when its neighborhood is sparsely represented.
We normalize the weights as $v_i=u_i/W$ and use the equivalent mean-one
rescaling in the fit. At fixed positive bandwidth and list size, these
weights are strictly positive, permutation-invariant, and continuous in the
distances. The next proposition gives the change in effective mass when we
add an entry.

\begin{proposition}[Marginal mass at fixed semantic scale]\label{prop:dup}
Holding the kernel bandwidth fixed, adding a benchmark with similarities
$k_i=k(d(x,x_i))$ changes the unnormalized effective mass by
\[
 \Delta W=\frac{1}{1+\sum_i k_i}
 -\sum_i\frac{k_i}{\rho_i(\rho_i+k_i)}.
\]
Against one isolated benchmark at distance $d$,
$\Delta W=(1-k(d))/(1+k(d))$.
\end{proposition}

In particular, isolated co-located entries share one unit of unnormalized
column mass. We measure the resulting movement of fitted rankings in
\Cref{sec:release-stability}.
We set $h_{\mathrm{dens}}=0.25\,\operatorname{median}_{i<j}d_{ij}$ within each
training pool and vary the multiplier in sensitivity analyses
(\Cref{sec:app-insight-listings,sec:app-release-stability}).

\paragraph{Density over reported benchmarks.}
Under sparse reporting, a model receives less than its intended share of
column weight if it reports only part of a crowded neighborhood. This
constraint applies to any positive column-only weighting rule
(\Cref{sec:app-mask-constraint}). In the reporting-aware form, we evaluate
the same density over the benchmarks each model reports,
$\rho_{mi}=\sum_{j:\,s_{mj}\,\mathrm{observed}}k(d_{ij})$, and use
$v_{mi}\propto\rho_{mi}^{-1}$ in place of $v_i$ in \Cref{eq:theta}. The
two forms coincide when every model reports every benchmark.
\Cref{sec:app-mask-aware,sec:app-natural} evaluate this form.

\section{Experiments}\label{sec:experiments}

For \textbf{Q1 (evidence retention)}, we replace a curated suite with all
eligible columns at a fixed model population and estimator, then compare
held-out rank agreement. For \textbf{Q2 (influence control)}, we hold the
measurements fixed and compare how far rankings move under density and
equal weighting when we re-list observations.

We use the density-weighted, score-equated ability $\theta_m$ after six
fitting rounds as the \bobname{} ranking. For retention and mask transfer,
we compare the same global ordering with each held-out benchmark ranking
within a fit. We evaluate query-conditioned prediction separately in the
AA reference. All \corpusname{} analyses condition on reported scores. For
rank correlations, we first average observed scores and $\theta_m$ within
release groups. We then average correlations over seeds within targets,
over targets within families, and equally across benchmark families
(\Cref{tab:corpora,tab:protocol-map}). We report every declared comparison
and discuss their scope in \Cref{sec:limitations}.

\begin{table}[H]
\centering\small
\caption{Evaluation matrices. Configurations are separately listed
model settings, and observed cells are model--benchmark scores.}
\label{tab:corpora}
\setlength{\tabcolsep}{4pt}
\begin{tabular}{lrrrr}
\toprule
Corpus & Configurations & Benchmarks & Observed cells & Observed (\%) \\
\midrule
Artificial Analysis & 605 & 14 & 5,441 & 64.2 \\
\corpusname{} & 354 & 148 & 5,371 & 10.3 \\
\bottomrule
\end{tabular}

\end{table}

\begin{table}[H]
\centering\footnotesize
\setlength{\tabcolsep}{4pt}
\caption{Evaluation map. Retention panels pair estimators on identical fitting
rows within each panel. Re-listing scenarios measure sensitivity of fitted
rankings to controlled changes in the recorded columns.}
\label{tab:protocol-map}
\begin{tabular}{@{}>{\raggedright\arraybackslash}p{0.15\linewidth}>{\raggedright\arraybackslash}p{0.31\linewidth}>{\raggedright\arraybackslash}p{0.27\linewidth}>{\raggedright\arraybackslash}p{0.20\linewidth}@{}}
\toprule
Analysis & Evaluation population & Intervention & Outcome \\
\midrule
Q1, strict panel & \bobStrictTargets{} targets, \bobStrictFamilies{} families &
Full pool versus curated suite, $\geq3$ curated scores/model & Global $\theta$ held-out Spearman \\
Q1, broad panel & \bobLowTargets{} targets, \bobLowFamilies{} families &
Same contrast, $\geq1$ curated score/model & Same ranking metric \\
Metadata screen & \bobMetadataLowTargets{} targets in the broad panel &
Remove \bobMetadataExcluded{} columns with protocol conflicts & Same ranking metric \\
Q2, re-listing & \bobReleaseReviewedConfigs{} configurations, eight source columns &
Four complete or partial copies & Mean absolute rank displacement \\
Affine transfer & 127 configurations (strict-panel rule) & Same copy interventions &
Same displacement metric \\
BIG-bench & 30 configurations, 131 retention targets & Suite expansion; six-source rule validation & Global $\theta$ Spearman; rank displacement \\
AA reference & \bobAARetainedRows{} configurations, 14 targets &
Leave one benchmark out & Query-conditioned Level/Profile \\
\bottomrule
\end{tabular}
\end{table}

\subsection{Evidence retention on a fixed model population}
\label{sec:release-prediction}

\paragraph{Corpus and withholding.}
We compile \corpusname{} from \bobReleaseReviewedCells{} developer-reported
scores, documenting their provenance and normalizing them. The resulting
matrix covers \bobReleaseBenchmarks{} benchmarks and
\bobReleaseReviewedConfigs{} model configurations across
\bobReleaseReviewedGroups{} release groups, with
\bobReleaseReviewedObservedPercent{}\% coverage. During source review, we
correct documented scale and setting issues. We test sensitivity to remaining
within-column protocol conflicts with the metadata screen below.
To keep benchmarks with shared question sources or derivations together, we
assign whole source-lineage families to five outer blocks for withholding
under each of two fixed seeds. We also use family labels for inference and
for one metadata-assisted weighting control.
\Cref{sec:app-release-data} details eligibility.

\paragraph{Fixed-population design.}
To compare the curated suite with the full pool, we fit both on identical
model rows within each panel. We construct the curated suite from individual
scores on ten benchmarks listed in the historical AA Index v3.0
\citep{artificialanalysis2026methodology}. Expanding this suite to the full
pool changes both the amount and domain composition of the evidence while
holding the model population fixed.
After withholding the target families in each fold, we require at least
three remaining curated observations per model in the strict panel and at
least one in the broad panel. In both panels, we require three observations
per column and apply variance filters using only training data. The strict
panel yields \bobStrictTargets{} targets in \bobStrictFamilies{} families,
and the broad panel yields \bobLowTargets{} targets in \bobLowFamilies{}
families. Because the fitting populations differ, we pair comparisons within
each panel.

\begin{table}[t]
\centering\small
\caption{Evidence retention on fixed fitting populations. Entries are
family-mean held-out Spearman correlations at six equating rounds. Panels
differ in fitting population. $\Delta$ (Full $-$ Curated) measures retention, while within-column
comparisons assess complete estimators. Within the \bobname{} group,
$\to$ replaces density weighting. Paired effects and uncertainty are in
\Cref{tab:insight-factorial}. $\dagger$ uses source-lineage metadata
unavailable to the description-only weighting rule. $\ddagger$ uses
description-based complete linkage, with cluster count matched to rounded
soft effective mass (\Cref{sec:app-semantic-grouping}).}
\label{tab:insight-comparison}
\setlength{\tabcolsep}{4pt}
\begin{tabular}{lrrrrrr}
\toprule
& \multicolumn{3}{c}{\shortstack{Strict panel\\($\geq3$ curated/model)}} & \multicolumn{3}{c}{\shortstack{Broad panel\\($\geq1$ curated/model)}} \\
Method & Curated & Full & $\Delta$ & Curated & Full & $\Delta$ \\
\midrule
\multicolumn{7}{l}{\textit{Existing aggregation rules}} \\
\quad Uniform standardized averaging & 0.819 & 0.827 & +0.008 & 0.742 & 0.765 & +0.023 \\
\quad Mean-imputed PC1 & 0.772 & 0.752 & -0.019 & 0.624 & 0.652 & +0.028 \\
\quad Observed-cell rank-one factor & 0.819 & 0.830 & +0.012 & 0.721 & 0.779 & +0.058 \\
\quad Observed-cell affine factor & 0.823 & 0.849 & +0.026 & 0.764 & 0.815 & +0.051 \\
\midrule
\multicolumn{7}{l}{\textit{\bobname{} and ablations}} \\
\quad BoB & 0.804 & 0.839 & \textbf{+0.035} & 0.765 & 0.802 & \textbf{+0.036} \\
\quad Equal-weight equating & 0.804 & 0.840 & +0.037 & 0.765 & 0.799 & +0.034 \\
\quad density $\to$ semantic groups$^\ddagger$ & 0.804 & 0.829 & +0.026 & 0.765 & 0.798 & +0.033 \\
\midrule
\multicolumn{7}{l}{\textit{Metadata-assisted control}} \\
\quad density $\to$ source-family weights$^\dagger$ & 0.798 & 0.852 & +0.054 & 0.762 & 0.806 & +0.044 \\
\bottomrule
\end{tabular}

\end{table}

\paragraph{Additional evidence benefits equating.}
On these matched populations, the full pool raises \bobname{}'s
family-mean Spearman by \bobStrictRetention{} in the strict panel
(95\% CI [\bobStrictRetentionLo{}, \bobStrictRetentionHi{}], Holm
$p=\bobStrictRetentionHolm{}$) and by \bobLowRetention{} in the
broad panel (Holm $p=\bobLowRetentionHolm{}$,
\Cref{tab:insight-comparison}).
Equal-weight equating gains \bobStrictFlatRetention{} and
\bobLowFlatRetention{}, respectively (\Cref{tab:insight-factorial}).
The gains show that both equating estimators benefit from evidence beyond the
tested curated suite. The retention intervals for uniform standardized
averaging cross zero (\Cref{tab:insight-factorial}).

\paragraph{Predictive performance across estimators.}
For Q1, we hold the estimator fixed when comparing pools. To assess
estimator choice separately, we compare full-pool fits. In the broad panel,
held-out Spearman rises from \bobLowFullPC{} for mean-imputed PC1 to
\bobLowFullMasked{} for the observed-cell rank-one factor and
\bobLowFullAffine{} for the affine factor. Both successive gains survive
Holm correction. The affine factor's predictive point estimates exceed
\bobname{}'s in both panels. Its differences from equal-weight equating are
nonsignificant after correction (\Cref{tab:insight-estimators}).
The \bobname{} minus equal-weight equating contrasts are
\bobStrictDensityDelta{} in the strict panel and \bobLowDensityDelta{} in
the broad panel, and both intervals cross zero. Held-out prediction
distinguishes the two evidence pools within equating, with little separation
among density weighting, equal weighting, and the two further rules in
\Cref{sec:app-mask-aware}.
\Cref{sec:app-insights} gives objectives, coverage, numerical checks, and
48-round sensitivities.

\paragraph{Description-based hard grouping is a viable alternative.}
We also test a weighting rule that gives each column the reciprocal of its
semantic cluster size. The grouping rule yields positive full-minus-curated
effects in both panels. To compare the weighting rules on the full pool, we
match effective mass and fitting conventions. Under these controls, the
difference between \bobname{} and semantic-group weighting is
\bobSemanticStrictDelta{} [\bobSemanticStrictLo{}, \bobSemanticStrictHi{}]
in the strict panel and \bobSemanticLowDelta{}
[\bobSemanticLowLo{}, \bobSemanticLowHi{}] in the broad panel.
The marginal interval favors \bobname{} in the strict panel and spans zero
in the broad panel (\Cref{sec:app-semantic-grouping}).

\paragraph{Protocol comparability limits the retention conclusion.}
To assess sensitivity to protocol comparability, we apply a metadata screen
that removes \bobMetadataExcluded{} columns with unresolved
evaluation-setting differences. Under the broad panel rules, \bobname{}'s
retention estimate remains positive at \bobMetadataLowRetention{} across
\bobMetadataLowTargets{} targets, with a marginal 95\% CI
[\bobMetadataLowRetentionLo{}, \bobMetadataLowRetentionHi{}]. The contrast
is nonsignificant after Holm correction over twenty comparisons
($p=\bobMetadataLowHolm{}$). The strict panel rules leave only
\bobMetadataStrictTargets{} targets. The screened comparison combines
changes in protocol eligibility, suite composition, and fitting population
(\Cref{sec:app-review-metadata,sec:app-semantic-families}).

To hold the fitting population and evaluation targets fixed, we compare
training pools with and without the screen on the rows and
\bobMatchedProtocolTargets{} targets that remain eligible after screening.
Held-out Spearman rises from \bobMatchedOriginalCurated{} to
\bobMatchedRestrictedCurated{} for the curated pool and from
\bobMatchedOriginalFull{} to \bobMatchedRestrictedFull{} for the full pool.
The larger gain for the curated pool narrows the retention gap from
\bobMatchedOriginalGap{} to \bobMatchedRestrictedGap{}
(\Cref{tab:semantic-metadata-cells}).
Because the screen changes protocol eligibility and pool composition
together, these point estimates describe sensitivity to the screened
training pool as a whole.

\subsection{Re-listing sensitivity and reporting overlap}
\label{sec:release-stability}

We test repeated influence with complete copies, a favorable case with
identical measurements. \Cref{sec:app-ruler-case} examines naturally
related tests with distinct measurements.

\paragraph{Complete re-listings.}
For each of eight \corpusname{} benchmarks, we add four score-and-mask
copies and refit. Mean absolute rank displacement is
\bobListingUniform{} for uniform standardized averaging,
\bobListingFlat{} for equal-weight equating, and \bobListingDensity{} for
density weighting, a \bobListingReduction{}\% reduction from equal-weight
equating. With paraphrased descriptions, the mean is \bobListingParaphrase{}.
The largest displacements are \bobListingExactMax{} ranks for exact copies
and \bobListingParaphraseMax{} for paraphrases, so stability is approximate.
For the partial-reporting tests below, we fix bandwidth at its pre-copy
value. The exact-copy mean is then \bobPartialFullDensity{}, close to
the re-estimated-bandwidth result (\Cref{tab:insight-listings}).
At fixed bandwidth, semantic grouping yields
\bobSemanticExactGroup{} for exact copies and \bobSemanticParaphraseGroup{}
for paraphrases, compared with \bobname{}'s \bobSemanticExactDensity{} and
\bobSemanticParaphraseDensity{}. Mean displacement favors \bobname{} in
all six declared conditions, with mixed worst-case ordering
(\Cref{tab:semantic-listing}).

\begin{table}[t]
\centering\small
\caption{Ranking sensitivity under complete and partial re-listings.
Mean absolute rank displacement is averaged over all 354 \corpusname{}
models and then scenarios. Each scenario adds four copies. Coverage is
conditional on observing the original column, and all four copies share one
mask. The density bandwidth is fixed at its pre-copy value. Scenarios are
interventions, not independent sampling trials.}
\label{tab:partial-main}
\setlength{\tabcolsep}{4pt}
\begin{tabular}{lrrr}
\toprule
Copy coverage & Scenarios & \shortstack{Equal-weight\\equating} & BoB \\
\midrule
Complete & 8 & 1.948 & 0.625 \\
25\% & 24 & 0.684 & 0.761 \\
50\% & 24 & 1.203 & 0.706 \\
75\% & 24 & 1.609 & 0.657 \\
Existing-column pattern & 8 & 1.089 & 0.727 \\
\bottomrule
\end{tabular}

\end{table}

\paragraph{Reporting overlap bounds the column form.}
At 25\% copy coverage, mean rank displacement rises from
\bobPartialQuarterFlat{} under equal-weight equating to
\bobPartialQuarterDensity{} under density weighting, with larger
displacement in 11 of the 24 scenarios (\Cref{tab:partial-main}). The
density-weighted mean is lower at 50\% and 75\% coverage and under the
tested existing-column reporting pattern.

The reporting mask explains the reversal. An isolated original and $k$
identical copies each receive unnormalized mass $1/(k+1)$. A model
reporting the original and $r_m$ copies thus receives $(1+r_m)/(k+1)$.
Observed mass varies by model at fixed column mass, a constraint on
positive column-only rules (\Cref{sec:app-mask-constraint}).
Nearest-description pairs have median directional coverage
\bobOverlapNearestMedian{}\%, and
\bobOverlapNearestLow{} of 148 directions have coverage at most 25\%
(\Cref{sec:app-semantic-overlap}). Reporting-count strata and source-level
variation appear in \Cref{sec:app-external-reporting}.

\paragraph{Density over reported benchmarks removes the reversal.}
We account for the mask by computing density over each model's reported
benchmarks (\Cref{sec:weighting}). Rank displacement is
smaller than under equal-weight equating in all 88 nonlinear scenarios,
including 25\% coverage (0.151 against 0.684, \Cref{tab:mask-aware}). A
clone-robust column rule \citep{berriaud2026consensus} performs best under
complete copies and retains a small average advantage at 25\% coverage.
The strict-panel predictive interval for reporting-aware density against
equal-weight equating spans zero. Third-party re-measurement of thirteen
benchmarks raises their share of model weight from \bobNatShareEqualBase{}\%
to \bobNatShareEqualRemeasure{}\% under equal weighting and from
\bobNatShareModelBase{}\% to \bobNatShareModelShared{}\% under reporting-aware
density with shared descriptions (\Cref{sec:app-natural}).

\paragraph{Initialization and residual rank movement.}
Matching score-and-mask columns lets us collapse exact copies under the first
listing's description. We retain them to trace how weights enter the fit.
Exact weight sharing yields mean rank displacement of \bobListingIdentity{}
places. Initializing with those weights eliminates movement in all sixteen
cases. Density weighting also changes neighboring weights, so weighted
initialization reduces movement only partially (\Cref{sec:app-insight-listings}).

\paragraph{Independent measurements at fixed weight.}
We simulate measurements with known capability identities at fixed total
weight per capability (\Cref{sec:app-insight-simulation}). Estimation error
is unchanged by copies and falls with fresh independent measurements.

\paragraph{The stability effect transfers to affine fitting.}
For the affine comparison, we apply the strict-panel observation and
variance filters once to the unperturbed matrix, retaining 127 models and
131 columns without holding out targets. On this population, density
weighting reduces complete-copy mean absolute rank displacement from
\bobSupportedFullAffine{} to \bobSupportedFullDensityAffine{}. At 25\%
nominal coverage, it raises mean displacement from
\bobSupportedQuarterAffine{} to \bobSupportedQuarterDensityAffine{}
(\Cref{tab:review-supported}). Both estimators show the complete-copy
benefit and low-overlap reversal.

\subsection{Validation on a separate score matrix}
\label{sec:external-validation}

We test retention and re-listing on a BIG-bench score matrix previously
unused for method development or evaluation \citep{srivastava2023bigbench}.
The corpora share task lineage because \corpusname{} retains BBH and BBEH
and BIG-bench includes source tasks for BBH. We use a common zero-shot,
temperature-zero policy, yielding \bobExternalModels{} configurations
from three architecture families and \bobExternalTasks{} parent tasks
(\bobExternalCoverage{}\% observed). Each parent score averages a fixed
set of eligible leaves. In five source-block folds, we match fitting rows
across comparisons and evaluate \bobExternalTargets{} nonconstant targets.
For the primary suite, we sample ten tasks with domain balance and average
results over five fixed seeds (\Cref{sec:app-external}).

\begin{table}[t]\centering\small
\caption{External retention under a ten-task suite policy.
Values are source-block-mean Spearman of global $\theta$ on 30 configurations
and 131 targets, without release-group averaging. $\Delta$ is Full minus
Ten-task. Intervals resample source blocks after averaging suite seeds.}
\label{tab:external-retention}
\begin{tabular}{@{}lrrrl@{}}
\toprule
Estimator & Ten-task & Full & $\Delta$ & 95\% CI for $\Delta$ \\
\midrule
Equal-weight equating & 0.1663 & 0.2450 & +0.0787 & [+0.0539, +0.1036] \\
\bobname{} & 0.1664 & 0.2489 & +0.0825 & [+0.0581, +0.1066] \\
Semantic grouping & 0.1663 & 0.2486 & +0.0823 & [+0.0592, +0.1055] \\
\bottomrule
\end{tabular}

\end{table}

In these matched comparisons, the full pool improves held-out Spearman
under both \bobname{} and equal-weight equating (Holm
$p=\bobExternalRetentionHolm{}$ for each). The full-pool density-minus-equal
contrast is nonsignificant (\bobExternalDensityDelta{}, 95\% CI
[\bobExternalDensityLo{}, \bobExternalDensityHi{}]). Retention intervals
are also positive against random suites, domain-balanced suites of 25 or 50
tasks, and the eligible BIG-bench Lite subset (\Cref{tab:external-policies}).
Domain quotas approximate fixed composition, and we record each suite's
actual allocation.

\paragraph{Observable conditions and semantic scope.}
To identify when column weighting reduces displacement, we test the rule
fixed from the earlier reporting analysis. It requires at least 50\%
conditional reporting overlap on every strong semantic-neighbor edge and
at least ten units of observed density mass at the model-level tenth
percentile. On six held-out intervention sources across 30 configurations,
it selects \bobExternalRuleSelected{} of \bobExternalRuleTotal{} scenarios.
In those scenarios, density weighting lowers mean displacement from
0.221 to 0.056 and meets the declared 0.01-Spearman noninferiority margin.
Displacement rises in \bobExternalRuleHarm{} individual scenarios
(\Cref{sec:app-external-rule}). Unperturbed \corpusname{} falls below both
thresholds (\Cref{sec:app-external-rule}). For sparse panels, we test the
reporting-aware form for multiplicity robustness
(\Cref{sec:app-mask-aware}).

Within parent tasks, item-reuse retrieval exceeds prevalence. The
cross-fitted residual correlation's association with description similarity
is \bobExternalResidualAssociation{}
[\bobExternalResidualLo{}, \bobExternalResidualHi{}].
Stronger outcome-dependent tilts reduce utility under column density and
equal-weight equating. Prediction also falls under both rules when we
transfer \corpusname{} reporting masks to BIG-bench without using scores.
The density-minus-equal interval spans zero
(\Cref{sec:app-real-masks}).

\subsection{Controlled mechanism reference on Artificial Analysis}
\label{sec:experiments-heldout}

We use the denser AA snapshot of 2026-09-03 to examine mechanisms with
less missingness (\bobAARetainedRows{} configurations, 14 benchmarks,
64\% observed cells). In fourteen folds, we predict each withheld
benchmark from the remaining columns. We evaluate \bobsupport{}, the task-conditioned extension in
Appendix~\ref{sec:app-task-extension}, with two metrics. \emph{Level} is
held-out Spearman correlation. \emph{Profile} correlates predicted and
observed ranks after removing a quadratic trend in the model's mean
observed training-benchmark percentile. This percentile defines the
reference for relative performance, separate from fitted $\theta_m$
(\Cref{sec:app-tests}). With fully nested
parameter selection, Level reaches \bobAggregationLevelThree{} and
Profile reaches \bobAggregationProfileThree{} for \bobsupport{}
(\Cref{sec:app-aggregation,sec:app-sensitivity}).

Removing density weights in the matched fit changes Profile from
\bobHeadlineProfileThree{} to \bobFlatProfileThree{}
($p=\bobFlatProfilePThree{}$) and raises the re-listing ability shift from
\bobCopyShiftThree{} to \bobFlatCopyShiftThree{} pre-copy standard
deviations (\Cref{tab:ablation}). We take the maximum absolute change in
$\theta$ across models in each of 14 four-copy scenarios and average the
standardized shifts. To test semantic placement, we reassign descriptions
at the original selected bandwidth in 200 reruns. Mean and maximum
Profile are \bobShuffleMeanThree{} and \bobShuffleMaxThree{}. The shuffle
comparison supports semantic placement in the full procedure. The
re-listings support density weighting for multiplicity robustness, with a
nonsignificant predictive increment.

We use the published AA Intelligence Index unchanged across folds as a
descriptive comparator. Its Profile difference from \bobsupport{} is
nonsignificant (Holm $p=\bobAAIndexProfileHolm{}$). \Cref{sec:app-aa} gives
the complete AA controls and \cref{tab:headline}.

\section{Interpretation and limitations}\label{sec:limitations}

\paragraph{Evidence retention and listing sensitivity.}
Suite expansion improves held-out ranking under both equating rules on
\corpusname{} and BIG-bench. Affine factors yield higher predictive point
estimates. Density weighting's incremental prediction intervals cross
zero. At substantial reporting overlap, density weighting reduces
re-listing sensitivity, including for paraphrases and supported affine
fits. Relative to semantic grouping, mean displacement falls and
worst-case comparisons are mixed.

\paragraph{Protocol comparability and reporting coverage.}
The metadata-restricted retention estimate is positive and nonsignificant
after correction. Provenance makes scores traceable. Differences in item
sets, metrics, and settings persist after equating. We evaluate reported
scores from eligible release groups. Unreported outcomes may follow a
different distribution. In the external stress test, stronger score-dependent
tilts lower utility under both estimators relative to matched random masks.
The actual \corpusname{} reporting process remains unidentified. At fixed
column mass, each model's observed weight depends on its mask. The column
form's stability reverses at 25\% copy coverage. During the discussion
period, we developed the reporting-aware form and tested it on exact
re-listings and two natural cases. Natural semantic neighbors can have low
reporting overlap and distinct scores and items.

\paragraph{Analysis and measurement scope.}
We designed the fixed-population study after inspecting earlier outcomes.
Suite expansion changes evidence amount and composition. We apply Holm
correction to declared comparisons, excluding the result-motivated
analysis choices. Intervals describe variation across benchmark families.
Model abilities are point estimates. We specified the external evaluation
before the new comparisons and documented support-only and data-conflict
corrections. Source-fold and intervention validations reuse one corpus
with three historical architecture families and one zero-shot protocol.
Across six held-out sources, the overlap rule identifies average re-listing
stability benefits with individual exceptions. Its mass threshold is
inactive, and sparse \corpusname{} fails the rule.

\paragraph{Semantic representation.}
Global held-out correlation evaluates one shared ordering and can be high
when general ability dominates and regional strengths go undetected. Query
correction and Profile assess task-specific departures using a training-pool
reference (\Cref{sec:app-tests}). We set weights using benchmark
descriptions, the tested Qwen3-Embedding encoders, and a density scale.
Item-reuse retrieval exceeds prevalence within parents. The cross-fitted
residual association interval spans zero. The weights set the emphasis on
similarly described benchmarks, with functional redundancy among heavily
downweighted benchmarks unverified. The kernel resolves near-identical
listings at a scale finer than broad topical balance (\Cref{sec:app-natural}).
Perplexity signatures offer a score-informed relation \citep{wu2026overlaps}.
Distinct-input additions yield small, mixed-sign predictive increments in
the tested dense pools.

\section{Conclusion}\label{sec:conclusion}

\bobname{} places eligible scores on a shared ability scale and limits the
influence of repeated listings. With the fitting population fixed, the full
pool improves held-out ranking under both equating rules. Column density
reduces rank displacement at substantial reporting overlap. At low overlap,
this benefit reverses in equating and affine fits. Computing density over
each model's reported benchmarks removes the reversal in controlled
re-listings. Transferring observed reporting masks to BIG-bench lowers
held-out agreement under column density and equal-weight equating. These
results support retaining eligible scores and weighting benchmarks within
each model's reported set.

\label{endofbody}

\clearpage
\subsubsection*{Reproducibility statement}
The study archive contains frozen scores, source records, setting
mappings, text inputs, embeddings, splits, and results. We generate
numerical prose and tables from saved outputs. The predictions, selected
parameters, and shortlists support reconstruction of rankings and paired
comparisons. We test target isolation, representative selection, support
matching, and agreement with the canonical fit. The records distinguish
the original evaluation from
exploratory extensions and identify fitting populations, stopping
settings, failed fits, and perturbation scenarios. We retain undefined
predictions from failed raw-score affine fits. The appendix specifies the
procedures, and the study archive records the score corpus,
provenance, and experimental texts.
For BIG-bench, we also record source and encoder revisions, evaluation
settings and job specifications, input-reuse labels, all 2,447 fitting jobs,
and documented deviations. Tests verify mask budgets, target isolation,
observable-only adoption features, and agreement with the canonical
estimator.
For the later analysis of observed masks and natural cases, we supply all
984 fits, sampled reporting masks, and score-blind task correspondence. An
independent audit reconstructs 27,488 correlations and all weights and
repeats eight canonical fits. The protocol documents the sensitivity
analysis and its reuse of the initial external validation corpus.

\subsubsection*{The use of large language models}
We evaluate language models. We also used them to assist benchmark-record
extraction, description and paraphrase construction, demonstration-query
canonicalization, writing, and code development. They assisted refinement
of experimental protocols and implementation of the follow-up comparisons.
For the experiments, we use frozen text and embedding inputs.
Deterministic analysis and generation scripts produce numerical results
from the retained score matrices. Parity tests and independent reaggregation
check the follow-up computations.
\Cref{sec:app-release-data,sec:app-text-construction} describe provenance.
The study archive preserves all frozen texts and experimental
evidence. We are responsible for the paper's content and claims.

\label{endofmaintext}

\bibliography{refs}

@article{liang2023holistic,
  title     = {Holistic Evaluation of Language Models},
  author    = {Percy Liang and Rishi Bommasani and Tony Lee and Dimitris Tsipras and
               Dilara Soylu and Michihiro Yasunaga and Yian Zhang and Deepak Narayanan and
               Yuhuai Wu and Ananya Kumar and Benjamin Newman and Binhang Yuan and
               Bobby Yan and Ce Zhang and Christian Alexander Cosgrove and Christopher D. Manning and
               Christopher R{\'e} and Diana Acosta-Navas and Drew Arad Hudson and
               Eric Zelikman and Esin Durmus and Faisal Ladhak and Frieda Rong and
               Hongyu Ren and Huaxiu Yao and Jue Wang and Keshav Santhanam and
               Laurel Orr and Lucia Zheng and Mert Yuksekgonul and Mirac Suzgun and
               Nathan Kim and Neel Guha and Niladri S. Chatterji and Omar Khattab and
               Peter Henderson and Qian Huang and Ryan Andrew Chi and Sang Michael Xie and
               Shibani Santurkar and Surya Ganguli and Tatsunori Hashimoto and
               Thomas Icard and Tianyi Zhang and Vishrav Chaudhary and William Wang and
               Xuechen Li and Yifan Mai and Yuhui Zhang and Yuta Koreeda},
  journal   = {Transactions on Machine Learning Research},
  year      = {2023},
  issn      = {2835-8856}
}

@article{srivastava2023bigbench,
  title     = {Beyond the Imitation Game: Quantifying and Extrapolating the Capabilities of
               Language Models},
  author    = {Aarohi Srivastava and Abhinav Rastogi and Abhishek Rao and
               Abu Awal Md Shoeb and Abubakar Abid and Adam Fisch and Adam R. Brown and
               Adam Santoro and Aditya Gupta and others},
  journal   = {Transactions on Machine Learning Research},
  year      = {2023},
  issn      = {2835-8856}
}

@inproceedings{wang2024mmlupro,
  title     = {{MMLU-Pro}: A More Robust and Challenging Multi-Task Language Understanding
               Benchmark},
  author    = {Yubo Wang and Xueguang Ma and Ge Zhang and Yuansheng Ni and
               Abhranil Chandra and Shiguang Guo and Weiming Ren and Aaran Arulraj and
               Xuan He and Ziyan Jiang and Tianle Li and Max Ku and Kai Wang and
               Alex Zhuang and Rongqi Fan and Xiang Yue and Wenhu Chen},
  booktitle = {Advances in Neural Information Processing Systems 37 (NeurIPS),
               Datasets and Benchmarks Track},
  year      = {2024}
}

@inproceedings{rein2024gpqa,
  title     = {{GPQA}: A Graduate-Level {Google-Proof} {Q}\&{A} Benchmark},
  author    = {David Rein and Betty Li Hou and Asa Cooper Stickland and Jackson Petty and
               Richard Yuanzhe Pang and Julien Dirani and Julian Michael and
               Samuel R. Bowman},
  booktitle = {Conference on Language Modeling (COLM)},
  year      = {2024}
}

@inproceedings{lightman2024verify,
  title     = {Let's Verify Step by Step},
  author    = {Hunter Lightman and Vineet Kosaraju and Yuri Burda and
               Harrison Edwards and Bowen Baker and Teddy Lee and Jan Leike and
               John Schulman and Ilya Sutskever and Karl Cobbe},
  booktitle = {International Conference on Learning Representations (ICLR)},
  year      = {2024},
  note      = {Introduces the {MATH-500} subset of the {MATH} test set}
}

@inproceedings{jain2025livecodebench,
  title     = {{LiveCodeBench}: Holistic and Contamination Free Evaluation of Large Language
               Models for Code},
  author    = {Naman Jain and King Han and Alex Gu and Wen-Ding Li and Fanjia Yan and
               Tianjun Zhang and Sida I. Wang and Armando Solar-Lezama and Koushik Sen and
               Ion Stoica},
  booktitle = {International Conference on Learning Representations (ICLR)},
  year      = {2025}
}

@inproceedings{tian2024scicode,
  title     = {{SciCode}: A Research Coding Benchmark Curated by Scientists},
  author    = {Minyang Tian and Luyu Gao and Shizhuo Dylan Zhang and Xinan Chen and
               Cunwei Fan and Xuefei Guo and Roland Haas and Pan Ji and
               Kittithat Krongchon and Yao Li and Shengyan Liu and Di Luo and Yutao Ma and
               Hao Tong and Kha Trinh and Chenyu Tian and Zihan Wang and Bohao Wu and
               Yanyu Xiong and Shengzhu Yin and Minhui Zhu and Kilian Lieret and
               Yanxin Lu and Genglin Liu and Yufeng Du and Tianhua Tao and Ofir Press and
               Jamie Callan and Eliu Huerta and Hao Peng},
  booktitle = {Advances in Neural Information Processing Systems 37 (NeurIPS),
               Datasets and Benchmarks Track},
  year      = {2024}
}

@inproceedings{yao2025taubench,
  title     = {{$\tau$}-bench: A Benchmark for Tool-Agent-User Interaction in Real-World
               Domains},
  author    = {Shunyu Yao and Noah Shinn and Pedram Razavi and Karthik Narasimhan},
  booktitle = {International Conference on Learning Representations (ICLR)},
  year      = {2025}
}

@inproceedings{barres2025tau2,
  title     = {{$\tau^2$}-{B}ench: Evaluating Conversational Agents in a Dual-Control
               Environment},
  author    = {Victor Barres and Honghua Dong and Soham Ray and Xujie Si and
               Karthik R. Narasimhan},
  booktitle = {Proceedings of the 43rd International Conference on Machine Learning},
  series    = {Proceedings of Machine Learning Research},
  volume    = {306},
  pages     = {6736--6782},
  year      = {2026}
}

@inproceedings{merrill2026terminalbench,
  title     = {{Terminal-Bench}: Benchmarking Agents on Hard, Realistic Tasks in Command
               Line Interfaces},
  author    = {Mike A. Merrill and Alexander G. Shaw and Nicholas Carlini and others},
  booktitle = {International Conference on Learning Representations},
  year      = {2026},
  note      = {Introduces {Terminal-Bench} 2.0}
}

@inproceedings{pyatkin2025ifbench,
  title     = {Generalizing Verifiable Instruction Following},
  author    = {Valentina Pyatkin and Saumya Malik and Victoria Graf and Hamish Ivison and
               Shengyi Huang and Pradeep Dasigi and Nathan Lambert and
               Hannaneh Hajishirzi},
  booktitle = {Advances in Neural Information Processing Systems 38 (NeurIPS),
               Datasets and Benchmarks Track},
  year      = {2025},
  note      = {Introduces the {IFBench} benchmark; preprint arXiv:2507.02833}
}

@article{phan2026hle,
  title   = {A Benchmark of Expert-Level Academic Questions to Assess {AI} Capabilities},
  author  = {{Center for AI Safety} and {Scale AI} and {HLE Contributors Consortium}},
  journal = {Nature},
  volume  = {649},
  number  = {8099},
  pages   = {1139--1146},
  year    = {2026},
  doi     = {10.1038/s41586-025-09962-4},
  note    = {Introduces the Humanity's Last Exam (HLE) benchmark; preprint arXiv:2501.14249}
}

@inproceedings{chiang2024arena,
  title     = {{Chatbot Arena}: An Open Platform for Evaluating {LLM}s by Human Preference},
  author    = {Wei-Lin Chiang and Lianmin Zheng and Ying Sheng and
               Anastasios Nikolas Angelopoulos and Tianle Li and Dacheng Li and
               Banghua Zhu and Hao Zhang and Michael Jordan and Joseph E. Gonzalez and
               Ion Stoica},
  booktitle = {Proceedings of the 41st International Conference on Machine Learning (ICML)},
  series    = {Proceedings of Machine Learning Research},
  volume    = {235},
  pages     = {8359--8388},
  year      = {2024}
}

@inproceedings{singh2025leaderboardillusion,
  title     = {The Leaderboard Illusion},
  author    = {Shivalika Singh and Yiyang Nan and Alex Wang and Daniel D'Souza and
               Sayash Kapoor and Ahmet {\"U}st{\"u}n and Sanmi Koyejo and Yuntian Deng and
               Shayne Longpre and Noah A. Smith and Beyza Ermis and Marzieh Fadaee and
               Sara Hooker},
  booktitle = {Advances in Neural Information Processing Systems 38 (NeurIPS),
               Datasets and Benchmarks Track},
  pages     = {86910--86964},
  year      = {2025},
  doi       = {10.52202/085713-2620}
}

@misc{ho2025rosetta,
  title        = {A Rosetta Stone for {AI} Benchmarks},
  author       = {Anson Ho and Jean-Stanislas Denain and David Atanasov and
                  Samuel Albanie and Rohin Shah},
  year         = {2025},
  howpublished = {arXiv:2512.00193},
  note         = {Preprint}
}

@inproceedings{berriaud2026clone,
  title     = {Clone-Robust Weights in Metric Spaces: Handling Redundancy Bias in
               Benchmark Aggregation},
  author    = {Damien Berriaud and Roger Wattenhofer},
  booktitle = {Proceedings of the 25th International Conference on Autonomous Agents
               and Multiagent Systems (AAMAS)},
  pages     = {2270--2278},
  year      = {2026},
  doi       = {10.65109/MJWJ6521}
}

@misc{berriaud2026consensus,
  title        = {Breaking the Illusion of Artificial Consensus: Clone-Robust Weighting
                  in Arbitrary Metric Spaces},
  author       = {Damien Berriaud and Supanat Wangsutthitham and Roger Wattenhofer},
  year         = {2026},
  howpublished = {arXiv:2602.24024v2},
  note         = {Preprint}
}

@inproceedings{zhang2024diversity,
  title     = {Inherent Trade-Offs between Diversity and Stability in Multi-Task
               Benchmarks},
  author    = {Guanhua Zhang and Moritz Hardt},
  booktitle = {Proceedings of the 41st International Conference on Machine Learning
               (ICML)},
  series    = {Proceedings of Machine Learning Research},
  volume    = {235},
  pages     = {58984--59002},
  year      = {2024}
}

@inproceedings{polo2024tinybenchmarks,
  title     = {{tinyBenchmarks}: Evaluating {LLM}s with Fewer Examples},
  author    = {Felipe {Maia Polo} and Lucas Weber and Leshem Choshen and Yuekai Sun and
               Gongjun Xu and Mikhail Yurochkin},
  booktitle = {Proceedings of the 41st International Conference on Machine Learning (ICML)},
  series    = {Proceedings of Machine Learning Research},
  volume    = {235},
  pages     = {34303--34326},
  year      = {2024}
}

@inproceedings{kipnis2025metabench,
  title     = {{metabench}: A Sparse Benchmark of Reasoning and Knowledge in Large Language
               Models},
  author    = {Alex Kipnis and Konstantinos Voudouris and Luca M. Schulze Buschoff and
               Eric Schulz},
  booktitle = {International Conference on Learning Representations (ICLR)},
  year      = {2025}
}

@inproceedings{rodriguez2021evaluation,
  title     = {Evaluation Examples Are Not Equally Informative: How Should That Change {NLP}
               Leaderboards?},
  author    = {Pedro Rodriguez and Joe Barrow and Alexander Hoyle and John P. Lalor and
               Robin Jia and Jordan Boyd-Graber},
  booktitle = {Proceedings of the 59th Annual Meeting of the Association for Computational
               Linguistics and the 11th International Joint Conference on Natural Language
               Processing (Volume 1: Long Papers)},
  pages     = {4486--4503},
  publisher = {Association for Computational Linguistics},
  year      = {2021},
  doi       = {10.18653/v1/2021.acl-long.346}
}

@inproceedings{lalor2016irt,
  title     = {Building an Evaluation Scale Using Item Response Theory},
  author    = {John P. Lalor and Hao Wu and Hong Yu},
  booktitle = {Proceedings of the 2016 Conference on Empirical Methods in Natural Language
               Processing (EMNLP)},
  pages     = {648--657},
  address   = {Austin, Texas},
  publisher = {Association for Computational Linguistics},
  year      = {2016},
  doi       = {10.18653/v1/D16-1062}
}

@misc{perlitz2024agreement,
  title        = {Do These {LLM} Benchmarks Agree? Fixing Benchmark Evaluation with
                  {BenchBench}},
  author       = {Yotam Perlitz and Ariel Gera and Ofir Arviv and Asaf Yehudai and
                  Elron Bandel and Eyal Shnarch and Michal Shmueli-Scheuer and
                  Leshem Choshen},
  year         = {2024},
  howpublished = {arXiv:2407.13696},
  note         = {Preprint}
}

@inproceedings{ruan2024observational,
  title     = {Observational Scaling Laws and the Predictability of Language Model
               Performance},
  author    = {Yangjun Ruan and Chris J. Maddison and Tatsunori Hashimoto},
  booktitle = {Advances in Neural Information Processing Systems 37 (NeurIPS)},
  year      = {2024}
}

@misc{qwen3embedding2025,
  title        = {{Qwen3} Embedding: Advancing Text Embedding and Reranking Through
                  Foundation Models},
  author       = {Yanzhao Zhang and Mingxin Li and Dingkun Long and Xin Zhang and
                  Huan Lin and Baosong Yang and Pengjun Xie and An Yang and
                  Dayiheng Liu and Junyang Lin and Fei Huang and Jingren Zhou},
  year         = {2025},
  howpublished = {arXiv:2506.05176},
  note         = {Technical report}
}

@book{lord1968statistical,
  title     = {Statistical Theories of Mental Test Scores},
  author    = {Frederic M. Lord and Melvin R. Novick},
  publisher = {Addison-Wesley},
  address   = {Reading, MA},
  year      = {1968}
}

@book{embretson2000irt,
  title     = {Item Response Theory for Psychologists},
  author    = {Susan E. Embretson and Steven P. Reise},
  publisher = {Lawrence Erlbaum Associates},
  address   = {Mahwah, NJ},
  year      = {2000}
}

@inproceedings{colombo2022benchmarking,
  title     = {What Are the Best Systems? New Perspectives on {NLP} Benchmarking},
  author    = {Pierre Colombo and Nathan Noiry and Ekhine Irurozki and
               Stephan Cl{\'e}men{\c{c}}on},
  booktitle = {Advances in Neural Information Processing Systems 35 (NeurIPS)},
  year      = {2022}
}

@inproceedings{perlitz2024efficient,
  title     = {Efficient Benchmarking (of Language Models)},
  author    = {Yotam Perlitz and Elron Bandel and Ariel Gera and Ofir Arviv and
               Liat Ein-Dor and Eyal Shnarch and Noam Slonim and
               Michal Shmueli-Scheuer and Leshem Choshen},
  booktitle = {Proceedings of the 2024 Conference of the North American Chapter of the
               Association for Computational Linguistics: Human Language Technologies
               (Volume 1: Long Papers)},
  pages     = {2519--2536},
  publisher = {Association for Computational Linguistics},
  year      = {2024},
  doi       = {10.18653/v1/2024.naacl-long.139}
}

@inproceedings{vivek2024anchor,
  title     = {Anchor Points: Benchmarking Models with Much Fewer Examples},
  author    = {Rajan Vivek and Kawin Ethayarajh and Diyi Yang and Douwe Kiela},
  booktitle = {Proceedings of the 18th Conference of the European Chapter of the
               Association for Computational Linguistics (Volume 1: Long Papers)},
  pages     = {1576--1601},
  address   = {St. Julian's, Malta},
  publisher = {Association for Computational Linguistics},
  year      = {2024}
}

@misc{artificialanalysis2026,
  title        = {Artificial Analysis: Independent Analysis of {AI} Models and {API}
                  Providers},
  author       = {{Artificial Analysis}},
  year         = {2026},
  howpublished = {\url{https://artificialanalysis.ai/}},
  note         = {Accessed 2026-09-03}
}

@misc{artificialanalysis2026methodology,
  title        = {Artificial Analysis Intelligence Benchmarking Methodology},
  author       = {{Artificial Analysis}},
  year         = {2026},
  howpublished = {\url{https://artificialanalysis.ai/methodology/intelligence-benchmarking}},
  note         = {Accessed 2026-11-19}
}

@misc{benchscope2026,
  title        = {{BenchScope}: How Many Independent Signals Does Your Benchmark Provide?},
  author       = {Tommy Sha and Stella Zhao},
  year         = {2026},
  howpublished = {arXiv:2603.29357},
  note         = {Preprint}
}

@misc{yao2026coresets,
  title        = {Coresets Before Score Sets: Evaluation-Unsupervised Prompt Subset
                  Selection for {LLM} Benchmarks},
  author       = {Jihan Yao and Gantavya Bhatt and Arnav Das and Peter Jin and Ke Bao and
                  Qiaolin Yu and Khushi Bhardwaj and Chang Su and Jialei Wang and
                  Yikai Zhu and Sugam Devare and Damon Mosk-Aoyama and Zhen Dong and
                  Venkat Krishna Srinivasan and Yineng Zhang and Oleksii Kuchaiev and
                  Jiantao Jiao and Banghua Zhu and Jeff Bilmes},
  year         = {2026},
  howpublished = {arXiv:2607.09739},
  note         = {Preprint}
}

@misc{smola2026submodular,
  title        = {Submodular Benchmark Selection},
  author       = {Alexander Smola},
  year         = {2026},
  howpublished = {arXiv:2605.02209},
  note         = {Preprint}
}

@misc{hudson2026metabench,
  title        = {Meta-Benchmarks for Financial-Services {LLM} Evaluation},
  author       = {Blair Hudson},
  year         = {2026},
  howpublished = {arXiv:2607.01740},
  note         = {Preprint}
}

@inproceedings{zhuang2024embedllm,
  title     = {{EmbedLLM}: Learning Compact Representations of Large Language Models},
  author    = {Richard Zhuang and Tianhao Wu and Zhaojin Wen and Andrew Li and
               Jiantao Jiao and Kannan Ramchandran},
  booktitle = {International Conference on Learning Representations (ICLR)},
  year      = {2025}
}

@inproceedings{song2025irtrouter,
  title     = {{IRT-Router}: Effective and Interpretable Multi-{LLM} Routing via Item
               Response Theory},
  author    = {Wei Song and Zhenya Huang and Cheng Cheng and Weibo Gao and Bihan Xu and
               GuanHao Zhao and Fei Wang and Runze Wu},
  booktitle = {Proceedings of the 63rd Annual Meeting of the Association for Computational
               Linguistics (Volume 1: Long Papers)},
  pages     = {15629--15644},
  publisher = {Association for Computational Linguistics},
  year      = {2025},
  doi       = {10.18653/v1/2025.acl-long.761}
}

@misc{chen2025irtnet,
  title        = {Learning Compact Representations of {LLM} Abilities via Item Response
                  Theory},
  author       = {Jianhao Chen and Chenxu Wang and Gengrui Zhang and Peng Ye and Lei Bai and
                  Wei Hu and Yuzhong Qu and Shuyue Hu},
  year         = {2025},
  howpublished = {arXiv:2510.00844},
  note         = {Preprint}
}

@inproceedings{ge2026agentpsych,
  title     = {Agent Psychometrics: Task-Level Performance Prediction in Agentic Coding
               Benchmarks},
  author    = {Chris Ge and Daria Kryvosheieva and Daniel Fried and Uzay Girit and
               Kaivalya Hariharan},
  booktitle = {Conference on Language Modeling},
  year      = {2026}
}

@misc{park2025anticipatory,
  title        = {Anticipatory Evaluation of Language Models: Predicting {LLM} Benchmark
                  Scores from Descriptions},
  author       = {Jungsoo Park and Ethan Mendes and Gabriel Stanovsky and Alan Ritter},
  year         = {2026},
  howpublished = {arXiv:2509.20645v4},
  note         = {Preprint}
}

@misc{li2025knnrouter,
  title        = {Rethinking Predictive Modeling for {LLM} Routing: When Simple {kNN}
                  Beats Complex Learned Routers},
  author       = {Yang Li},
  year         = {2025},
  howpublished = {arXiv:2505.12601},
  note         = {Preprint}
}

@book{kolen2014equating,
  title     = {Test Equating, Scaling, and Linking: Methods and Practices},
  author    = {Michael J. Kolen and Robert L. Brennan},
  edition   = {3rd},
  publisher = {Springer},
  address   = {New York, NY},
  year      = {2014},
  doi       = {10.1007/978-1-4939-0317-7}
}

@article{stocking1983metric,
  title   = {Developing a Common Metric in Item Response Theory},
  author  = {Martha L. Stocking and Frederic M. Lord},
  journal = {Applied Psychological Measurement},
  volume  = {7},
  number  = {2},
  pages   = {201--210},
  year    = {1983},
  doi     = {10.1177/014662168300700208}
}

@inproceedings{shi2022mgsm,
  title = {Language Models are Multilingual Chain-of-Thought Reasoners},
  author = {Freda Shi and Mirac Suzgun and Markus Freitag and others},
  booktitle = {International Conference on Learning Representations (ICLR)},
  year = {2023}
}

@article{cassano2022multiple,
  title = {{MultiPL-E}: A Scalable and Polyglot Approach to Benchmarking Neural Code Generation},
  author = {Federico Cassano and John Gouwar and Daniel Nguyen and others},
  journal = {IEEE Transactions on Software Engineering},
  volume = {49},
  number = {7},
  pages = {3675--3691},
  year = {2023},
  doi = {10.1109/TSE.2023.3267446}
}

@article{he2024multiif,
  title = {{Multi-IF}: Benchmarking {LLMs} on Multi-Turn and Multilingual Instructions Following},
  author = {Yun He and Di Jin and Chaoqi Wang and others},
  journal = {arXiv preprint arXiv:2410.15553},
  year = {2024}
}

@inproceedings{wang2025polymath,
  title = {{PolyMath}: Evaluating Mathematical Reasoning in Multilingual Contexts},
  author = {Yiming Wang and Pei Zhang and Jialong Tang and others},
  booktitle = {Advances in Neural Information Processing Systems 38 (NeurIPS),
               Datasets and Benchmarks Track},
  year = {2025}
}

@inproceedings{duan2024vlmevalkit,
  title = {{VLMEvalKit}: An Open-Source {ToolKit} for Evaluating Large Multi-Modality Models},
  author = {Duan, Haodong and Yang, Junming and Qiao, Yuxuan and Fang, Xinyu and Chen, Lin and Liu, Yuan and Dong, Xiaoyi and Zang, Yuhang and Zhang, Pan and Wang, Jiaqi and Lin, Dahua and Chen, Kai},
  booktitle = {Proceedings of the 32nd ACM International Conference on Multimedia},
  year = {2024},
  pages = {11198--11201},
  doi = {10.1145/3664647.3685520},
  url = {https://arxiv.org/html/2407.11691v1#S3}
}

@inproceedings{chen2024mmstar,
  title = {Are We on the Right Way for Evaluating Large Vision-Language Models?},
  author = {Lin Chen and Jinsong Li and Xiaoyi Dong and others},
  booktitle = {Advances in Neural Information Processing Systems 37 (NeurIPS)},
  year = {2024}
}

@inproceedings{lu2023mathvista,
  title = {{MathVista}: Evaluating Mathematical Reasoning of Foundation Models in Visual Contexts},
  author = {Pan Lu and Hritik Bansal and Tony Xia and others},
  booktitle = {International Conference on Learning Representations (ICLR)},
  year = {2024}
}

@inproceedings{zou2024dynamath,
  title = {{DynaMath}: A Dynamic Visual Benchmark for Evaluating Mathematical Reasoning Robustness of Vision Language Models},
  author = {Chengke Zou and Xingang Guo and Rui Yang and Junyu Zhang and Bin Hu and Huan Zhang},
  booktitle = {International Conference on Learning Representations (ICLR)},
  year = {2025}
}

@inproceedings{wang2024muirbench,
  title = {{MuirBench}: A Comprehensive Benchmark for Robust Multi-image Understanding},
  author = {Fei Wang and Xingyu Fu and James Y. Huang and others},
  booktitle = {International Conference on Learning Representations (ICLR)},
  year = {2025}
}

@article{liu2024ocrbench,
  title = {{OCRBench}: On the Hidden Mystery of {OCR} in Large Multimodal Models},
  author = {Yuliang Liu and Zhang Li and Mingxin Huang and Biao Yang and Wenwen Yu and Chunyuan Li and Xu-Cheng Yin and Cheng-Lin Liu and Lianwen Jin and Xiang Bai},
  journal = {Science China Information Sciences},
  volume = {67},
  number = {12},
  pages = {220102},
  year = {2024},
  doi = {10.1007/s11432-024-4235-6}
}

@article{leinster2012diversity,
  title     = {Measuring Diversity: The Importance of Species Similarity},
  author    = {Tom Leinster and Christina A. Cobbold},
  journal   = {Ecology},
  volume    = {93},
  number    = {3},
  pages     = {477--489},
  year      = {2012},
  doi       = {10.1890/10-2402.1}
}

@misc{willerton2015spread,
  title         = {Spread: A Measure of the Size of Metric Spaces},
  author        = {Simon Willerton},
  year          = {2015},
  eprint        = {1209.2300},
  archivePrefix = {arXiv},
  primaryClass  = {math.MG},
  note          = {Version 2},
  url           = {https://arxiv.org/abs/1209.2300v2}
}

@inproceedings{himmi2024missing,
  title = {Towards More Robust {NLP} System Evaluation: Handling Missing Scores in Benchmarks},
  author = {Himmi, Anas and Irurozki, Ekhine and Noiry, Nathan and Cl{\'e}men{\c{c}}on, Stephan and Colombo, Pierre},
  booktitle = {Findings of the Association for Computational Linguistics: EMNLP 2024},
  year = {2024},
  pages = {11759--11785},
  publisher = {Association for Computational Linguistics},
  doi = {10.18653/v1/2024.findings-emnlp.688},
  url = {https://aclanthology.org/2024.findings-emnlp.688/}
}

@misc{uzunoglu2026flaw,
  title = {The Flaw of Averages: Measuring Benchmark-Level Distributional Robustness},
  author = {Arda Uzunoglu and Tianjian Li and Daniel Khashabi},
  year = {2026},
  eprint = {2509.25671},
  archivePrefix = {arXiv},
  note = {Version 2},
  url = {https://arxiv.org/abs/2509.25671v2}
}

@inproceedings{siska2024examining,
  title = {Examining the Robustness of {LLM} Evaluation to the Distributional Assumptions of Benchmarks},
  author = {Charlotte Siska and Katerina Marazopoulou and Melissa Ailem and James Bono},
  booktitle = {Proceedings of the 62nd Annual Meeting of the Association for Computational Linguistics (Volume 1: Long Papers)},
  year = {2024},
  pages = {10406--10421},
  doi = {10.18653/v1/2024.acl-long.560},
  url = {https://aclanthology.org/2024.acl-long.560/}
}

@inproceedings{wu2026overlaps,
  title = {Mapping Overlaps in Benchmarks through Perplexity in the Wild},
  author = {Siyang Wu and Honglin Bao and Sida Li and Ari Holtzman and James Evans},
  booktitle = {International Conference on Learning Representations},
  year = {2026},
  url = {https://proceedings.iclr.cc/paper_files/paper/2026/file/d69ec8cae53d88f97a9449fbef59ce5d-Paper-Conference.pdf}
}

@inproceedings{hsieh2024ruler,
  title = {{RULER}: What's the Real Context Size of Your Long-Context Language Models?},
  author = {Hsieh, Cheng-Ping and Sun, Simeng and Kriman, Samuel and Acharya, Shantanu and Rekesh, Dima and Jia, Fei and Zhang, Yang and Ginsburg, Boris},
  booktitle = {Conference on Language Modeling},
  year = {2024},
  url = {https://arxiv.org/abs/2404.06654}
}

@misc{projectapertus2025apertus,
  title = {{Apertus}: Democratizing Open and Compliant {LLMs} for Global Language Environments},
  author = {{Project Apertus} and others},
  year = {2025},
  eprint = {2509.14233},
  archivePrefix = {arXiv},
  url = {https://arxiv.org/abs/2509.14233v2}
}

@inproceedings{suzgun2023bigbenchhard,
  title = {Challenging {BIG}-Bench Tasks and Whether Chain-of-Thought Can Solve Them},
  author = {Suzgun, Mirac and Scales, Nathan and Schärli, Nathanael and Gehrmann, Sebastian and Tay, Yi and Chung, Hyung Won and Chowdhery, Aakanksha and Le, Quoc and Chi, Ed and Zhou, Denny and Wei, Jason},
  booktitle = {Findings of the Association for Computational Linguistics: ACL 2023},
  year = {2023},
  pages = {13003--13051},
  doi = {10.18653/v1/2023.findings-acl.824},
  url = {https://aclanthology.org/2023.findings-acl.824/}
}

@inproceedings{kazemi2025bbeh,
  title={{BIG}-Bench Extra Hard},
  author={Kazemi, Mehran and Fatemi, Bahare and Bansal, Hritik and Palowitch, John and Anastasiou, Chrysovalantis and Mehta, Sanket Vaibhav and Jain, Lalit K and Aglietti, Virginia and Jindal, Disha and Chen, Peter and Dikkala, Nishanth and Tyen, Gladys and Liu, Xin and Shalit, Uri and Chiappa, Silvia and Olszewska, Kate and Tay, Yi and Tran, Vinh Q. and Le, Quoc V and Firat, Orhan},
  booktitle={Proceedings of the 63rd Annual Meeting of the Association for Computational Linguistics (Volume 1: Long Papers)},
  year={2025},
  pages={26473--26501},
  doi={10.18653/v1/2025.acl-long.1285},
  url={https://aclanthology.org/2025.acl-long.1285/}
}

@inproceedings{balduzzi2018reevaluating,
  title     = {Re-evaluating Evaluation},
  author    = {David Balduzzi and Karl Tuyls and Julien Perolat and Thore Graepel},
  booktitle = {Advances in Neural Information Processing Systems (NeurIPS)},
  volume    = {31},
  pages     = {3272--3283},
  year      = {2018}
}

@inproceedings{liu2025reevaluating,
  title     = {Re-evaluating Open-ended Evaluation of Large Language Models},
  author    = {Siqi Liu and Ian Gemp and Luke Marris and Georgios Piliouras and
               Nicolas Heess and Marc Lanctot},
  booktitle = {International Conference on Learning Representations},
  year      = {2025}
}

@article{wilks1938weighting,
  title     = {Weighting Systems for Linear Functions of Correlated Variables When
               There Is No Dependent Variable},
  author    = {S. S. Wilks},
  journal   = {Psychometrika},
  volume    = {3},
  number    = {1},
  pages     = {23--40},
  year      = {1938},
  doi       = {10.1007/BF02287917}
}

@article{paruolo2013ratings,
  title     = {Ratings and Rankings: Voodoo or Science?},
  author    = {Paolo Paruolo and Michaela Saisana and Andrea Saltelli},
  journal   = {Journal of the Royal Statistical Society Series A: Statistics in Society},
  volume    = {176},
  number    = {3},
  pages     = {609--634},
  year      = {2013},
  doi       = {10.1111/j.1467-985X.2012.01059.x}
}

@misc{navasardyan2026audit,
  title         = {A Statistical Audit of Physical {AI} Benchmark Redundancy},
  author        = {Zaruhi Navasardyan and Hrant Davtyan},
  year          = {2026},
  eprint        = {2608.25940},
  archivePrefix = {arXiv},
  primaryClass  = {cs.RO}
}

@article{sanderson2017skill,
  title   = {Skill and Independence Weighting for Multi-Model Assessments},
  author  = {Benjamin M. Sanderson and Michael Wehner and Reto Knutti},
  journal = {Geoscientific Model Development},
  volume  = {10},
  number  = {6},
  pages   = {2379--2395},
  year    = {2017},
  doi     = {10.5194/gmd-10-2379-2017}
}

@book{nardo2008handbook,
  title     = {Handbook on Constructing Composite Indicators: Methodology and User Guide},
  author    = {Michela Nardo and Michaela Saisana and Andrea Saltelli and Stefano Tarantola and Anders Hoffmann and Enrico Giovannini},
  publisher = {OECD Publishing},
  address   = {Paris},
  year      = {2008},
  doi       = {10.1787/9789264043466-en}
}
\bibliographystyle{iclr2027_conference}

\appendix
\clearpage
\raggedbottom
\section*{Appendix guide}
\label{startofappendix}
\textbf{How do we combine scores?} \Cref{sec:app-method} explains how we fit
abilities and compute weights, then derives what happens to the weights
under re-listing. \Cref{sec:app-release-data} describes how we construct
and evaluate \corpusname{}.

\textbf{What do additional scores contribute?}
\Cref{sec:app-insights,sec:app-review,sec:app-semantic} trace how retention
and re-listing results depend on the fitting population, protocol screen,
and reporting overlap. \Cref{sec:app-external} tests these results on
BIG-bench, compares description similarity with input reuse, and examines
reporting patterns and real re-measurements.

\textbf{What can the scores tell us about a particular task or model pair?}
\Cref{sec:app-task-extension,sec:app-release-results,sec:app-aa} describe
the task-conditioned extension, test its sensitivity, and give the
Artificial Analysis controls. \Cref{sec:app-close-comparisons} examines
closely ranked checkpoints and estimates HMMT scores from nearby and
distant benchmarks.

The arXiv ancillary files include the \corpusname{} scores, provenance,
benchmark descriptions, and \emph{Appendix Details}. The study archive
preserves the full experimental evidence. In the original numbering, Tables~28,
52 and 54 correspond to \cref{tab:natural-signal,tab:external-policies,tab:external-selective}, respectively.

\section{Additional method details}\label{sec:app-method}

\subsection{Fitting details}\label{sec:app-impl}

\begin{table}[H]
\centering\small
\caption{Global \bobname{} fitting procedure. The canonical estimator stops
after six rounds. Each training pool determines its score matrix and geometry.}
\label{tab:fit-procedure}
\begin{minipage}{0.96\linewidth}
\hrule\smallskip
\textbf{Input:} eligible normalized scores $S$, observation mask $O$,
description distances $D$, and number of rounds $T=6$.
\begin{enumerate}
\setlength{\itemsep}{3pt}
\setlength{\parsep}{0pt}
\item Compute $h_{\mathrm{dens}}=0.25\,\operatorname{median}_{i<j}d_{ij}$,
$u_i=1/\sum_j\exp[-d_{ij}^2/(2h_{\mathrm{dens}}^2)]$, and
$v_i=N u_i/\sum_j u_j$ for the $N$ retained columns.
\item Initialize each $\theta_m$ by the mean midpoint percentile of its
observed scores, then standardize across models. Set $a_i=2$, $b_i=0$,
and $w_{mi}=O_{mi}$.
\item For each round $t=1,\ldots,T$, perform the following updates in order.
\begin{enumerate}
\setlength{\itemsep}{2pt}
\item Fit each curve by minimizing
$\sum_{m:O_{mi}=1}w_{mi}[s_{mi}-\Phi(a_i(\theta_m-b_i))]^2$
with L-BFGS-B over $\log a_i\in[\log0.2,\log20]$ and $b_i\in[-4,4]$.
\item Minimize \Cref{eq:theta} for each model on the grid
$[-3.5,3.5)$ with step 0.02. At an interior minimum, refine with the
three-point parabola, limiting the offset to one grid step. A curvature
denominator of magnitude at most $10^{-12}$ gives zero offset.
\item Standardize abilities and transport curves jointly:
$\theta\leftarrow(\theta-\mu)/s$, $a\leftarrow sa$,
$b\leftarrow(b-\mu)/s$.
\item Recompute raw residuals $e_{mi}=s_{mi}-f_i(\theta_m)$ and
$\sigma_i=\max(\operatorname{sd}_{m:O_{mi}=1}e_{mi},10^{-4})$,
using the population standard deviation. For $t\geq2$, update observed-cell
weights to $\max(10^{-3},\min(1,(3/|e_{mi}/\sigma_i|)^2))$.
Missing cells retain zero weight.
\end{enumerate}
\item Return $\theta,a,b$, residuals, residual scales, and the final weights.
Robust updates after a round enter the next round's fit. The last update
also supplies weights for the optional residual correction.
\end{enumerate}
\smallskip\hrule
\end{minipage}
\end{table}

We assign average ranks to ties at initialization and apply the
curve-optimization bounds before the joint standardization in
\Cref{tab:fit-procedure}. The transformation preserves $a_i(\theta_m-b_i)$
and thus every expected score.

\subsection{Inverse-density weighting and duplication decomposition}\label{sec:app-proof}

We fix the benchmark descriptions and kernel bandwidth before adding $x$,
and write $\rho_i=\sum_j k(d_{ij})$ and $k_i=k(d(x,x_i))$. Each existing
density then becomes $\rho_i+k_i$, and the new point has density
$\rho_x=1+\sum_i k_i$. The resulting change in effective mass is
\[
  \Delta W
  = \frac{1}{1+\sum_i k_i}
    + \sum_i\left(\frac{1}{\rho_i+k_i}-\frac{1}{\rho_i}\right)
  = \frac{1}{1+\sum_i k_i}
    - \sum_i\frac{k_i}{\rho_i(\rho_i+k_i)} .
\]
$\Delta W$ decreases in each $k_i$ at fixed original densities $\rho_i$.
Re-estimating the bandwidth can change those densities, an effect we test
separately in \Cref{sec:app-insight-listings}.
When every $k_i=0$, $\Delta W=1$. If $x$ coincides with a member of an isolated
co-located group $G$ of size $n_G$, then $k_i=1$ and $\rho_i=n_G$ for
$i\in G$, giving
\[
  \Delta W=\frac{1}{1+n_G}
  -\frac{n_G}{n_G(n_G+1)}=0.
\]
Against one isolated benchmark at distance $d$, $\rho_1=1$ and
$k_1=k(d)$ yield $\Delta W=(1-k)/(1+k)$.

Exact cancellation assumes an isolated group. With neighbors present, a
copy raises their densities and can make $\Delta W$ slightly negative.
The unnormalized mass of $n_G$ co-located benchmarks is
\[
  \frac{n_G}{n_G+c_G},
  \qquad
  c_G=\sum_{j\notin G} k\big(d(x_G,x_j)\big).
\]
This mass differs from the single-copy value $1/(1+c_G)$ by at most $c_G$.
We treat each original benchmark as a singleton before copying. On AA,
$c_G$ ranges from \bobNeighborMassMinThree{} to \bobNeighborMassMaxThree{}
outside the two close pairs, and from \bobAimeNeighborMassMinThree{} to
\bobAimeNeighborMassMaxThree{} for the two AIME entries.

\paragraph{Relation to similarity-sensitive diversity.}
Let $Z_{ij}=k(d_{ij})$ and $p_i=1/N$. The order-zero diversity of
\citet{leinster2012diversity} is
\[
 D_0^Z(p)=\sum_i\frac{p_i}{(Zp)_i}.
\]
For uniform $p$, $(Zp)_i=\rho_i/N$, so
$D_0^Z(p)=\sum_i1/\rho_i=W$. We use the individual inverse-density terms
of this diversity measure as benchmark weights. The identity concerns
effective mass. The comparison below measures rank changes after refitting.

\paragraph{Why the fitted ranking still moves.}
With four added copies, we separate the effects of holding group weight
fixed, restoring inverse-density weights, and refitting. Rankings are
unchanged when both group weight and item parameters stay fixed. With item
parameters still fixed, the shared-neighbor $O(c_G)$ term in inverse-density
weights yields a scenario-averaged maximum displacement of
\bobCopyFixedRankOne{} ranks. Refitting curves and robust cell weights
raises that maximum to \bobCopyFullRankOne{} ranks. The scenario-averaged
maximum change in $\theta$ is \bobCopyShiftThree{} standard deviations,
and the median gap between adjacent models is \bobAdjacentGapFour{}.
Appendix Details contains the effective-mass plots and the Kendall $\tau$
results for the fourteen exact-copy scenarios (Figures~3 and~4).

\subsection{Semantic and score channels}\label{sec:app-channels}

\begin{figure}[H]
\centering
\includegraphics[width=\linewidth]{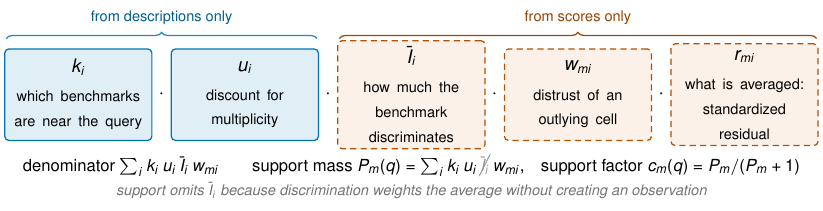}
\caption{\textbf{Semantic and score factors retain distinct roles in the
task-conditioned field.} The query kernel and inverse density come from
descriptions. Information, robust influence, and residuals come from scores.
Information weights direct the residual average. Support mass counts
relevant observed evidence through $g_{mi}(q)w_{mi}$.}
\label{fig:factors}
\end{figure}

\Cref{fig:pipeline} separates the semantic and score channels of the
estimator. \Cref{fig:saturation} illustrates how the score channel equates
difficulty ranges. \Cref{tab:touchpoints} summarizes where scores enter,
and \Cref{fig:factors} shows how the two channels contribute to the
task-conditioned field.

\subsubsection{Score weights and information}\label{sec:app-score-weights}

For observed cells, we use the robust weight
$w_{mi}=\max(10^{-3},\min(1,(3/|r_{mi}|)^2))$ and assign zero to missing
cells. We update these weights after the second and subsequent
rounds for use in the next fit. The rule gives unit weight to residuals
within three standard deviations and downweights larger residuals
quadratically. These weights also enter the residual field and support mass.

To weight the residual field by precision, we use mean information
\begin{equation}
  \bar I_i=\frac{1}{M}\sum_{m=1}^{M}
  \frac{\big(a_i\phi(a_i(\theta_m-b_i))\big)^2}
       {\max(\sigma_i^2,10^{-6})},
\end{equation}
where $\phi$ is the standard normal density. This quantity measures how
strongly benchmark $i$ separates the fitted model population relative to
its residual scale. Its role is confined to precision weighting in the
residual field.

\begin{table}[H]
\centering
\small
\caption{Uses of model scores in the global fit and optional task extension.
Information weighting and query-bandwidth selection belong to the extension.
Semantic weights are fixed from descriptions before score-side quantities enter.}
\label{tab:touchpoints}
\label{sec:app-touchpoints}
\begin{tabular}{p{0.29\linewidth}p{0.25\linewidth}p{0.36\linewidth}}
\toprule
Use of scores & Enters as & Role \\
\midrule
Characteristic-curve fit & $a_i,b_i,\theta_m$ in
\Cref{eq:icc,eq:theta} & equates scales and supplies residuals \\
Per-benchmark residual scale & $\sigma_i$ in $r_{mi}$ and $\bar I_i$ &
empirical residual standard deviation with numerical floor \\
Mean item information & $\bar I_i$ in \Cref{eq:field} &
precision-weights the local residual average \\
Population location and scale & standardization of $s_{mi}$ and $\theta_m$ &
sets location and scale units \\
Robust cell weight & $w_{mi}$ in \Cref{eq:theta,eq:field} &
limits the influence of one cell's residual \\
Cross-validation of $h$ & field smoothness &
evaluation-side selection with a nested check \\
\bottomrule
\end{tabular}
\end{table}

\subsection{Task-relevance controls}\label{sec:app-relevance-controls}

We vary $g$ while holding equated scores, residuals, information, robust
weights, and the form of support regularization fixed. Hard rules assign
unit relevance within a selected category, neighborhood, or semantic
cluster. The all-benchmark rule assigns it to every observation. The
support-matched top-$k$ rule matches the continuous rule's observed support
(\Cref{sec:app-release-controls}). Across these rules, $g$ also determines
correction strength through $P_m^{(g)}(q)$.

\subsection{Frozen text construction and embedding}%
\label{sec:app-text-construction}

On 2026-08-29, we generated the fourteen English AA benchmark descriptions
and their paraphrases with \texttt{claude-fable-5}. For each description,
we requested a plain-text paragraph of 2--4 sentences and roughly 40--80 words
covering the domain, abilities, item format, and answer language. For each
paraphrase, we retained meaning and length and varied wording and sentence
structure. We canonicalized each of the five demonstration queries once to
match the descriptions in length and facets, then fixed all texts before
analysis. The study archive includes the original and canonical queries.

We embed all text with \texttt{Qwen/Qwen3-Embedding-8B} through
SentenceTransformers. We run the encoder in float16 on an NVIDIA L40S
with maximum sequence length 1024 and batch size 8, then store normalized
float32 embeddings. We encode both benchmark descriptions and task queries
as documents with an empty instruction prefix. The geometry uses the
normalized vectors directly, with no centering or component removal.

The study archive provides the \corpusname{} and AA descriptions, paraphrases,
frozen vectors and score-column identifiers. The benchmark publications
describe their evaluations
\citep{rein2024gpqa,phan2026hle,wang2024mmlupro,lightman2024verify,jain2025livecodebench,tian2024scicode,pyatkin2025ifbench,yao2025taubench,barres2025tau2,merrill2026terminalbench}.

\subsection{Relation to perplexity-derived benchmark signatures}
\label{sec:app-signatures}

\citet{wu2026overlaps} estimate functional overlap by selecting corpus tokens
with a reference panel and comparing mean within-model standardized
perplexity profiles by Spearman correlation. We use benchmark relationships
to allocate influence when pooling scores. Retaining related measurements
can widen coverage, and reporting patterns determine which measurements
contribute to each model's estimate. \Cref{tab:signature-comparison}
compares the inputs and outputs of the two constructions.

\begin{table}[H]
\centering\small
\caption{Benchmark signatures and \bobname{} address related stages of
model evaluation. Entries compare the inputs and outputs of the published
constructions.}
\label{tab:signature-comparison}
\begin{tabular}{@{}>{\raggedright\arraybackslash}p{0.20\linewidth}>{\raggedright\arraybackslash}p{0.36\linewidth}>{\raggedright\arraybackslash}p{0.37\linewidth}@{}}
\toprule
Property & Benchmark signatures \citep{wu2026overlaps} & \bobname{} \\
\midrule
Overlap signal & Token-perplexity profiles selected to predict scores &
Proximity of benchmark descriptions \\
Geometry inputs & Reference-model benchmark scores and corpus token probabilities &
Descriptions and a fixed text encoder \\
Reported output & Pairwise benchmark-overlap map and performance regressions &
Benchmark coefficients and equated model abilities \\
Population dependence & Re-estimating signatures can change the overlap map &
Fixed-list coefficients stay fixed, fitted abilities depend on scores and masks \\
Question studied & How benchmarks share predictive signatures &
How retained observations and their weights affect model comparison \\
\bottomrule
\end{tabular}
\end{table}

One possible substitution uses signature correlations from a common
reference panel to supply angular distances and a kernel $K$ for the
same weights $u_i=(\sum_j K_{ij})^{-1}$. The signature map's use in a new
population would require transfer of the learned relationships.
Task-conditioned use would additionally require query-to-benchmark
similarities. This substitution remains untested.

For the tested description geometry, the association with cross-fitted
residual correlation is \bobExternalResidualAssociation{}
[\bobExternalResidualLo{}, \bobExternalResidualHi{}], an interval spanning
zero (\Cref{sec:app-external-semantics}).

\paragraph{Future content representations.}\label{sec:future-content}
Subset or item representations could distinguish scoring composition,
measurement precision, and semantic coverage while retaining aggregate
input scores. Question embeddings could allocate shared weight according to
question-content overlap. This approach requires access to question text,
which is public for only some benchmarks, as well as computation to embed
items and normalization across item sets of different sizes. Estimating regional performance from
aggregate scores would also require assumptions about which aspects of
performance those scores identify.

\section{\corpusname{} construction and evaluation}\label{sec:app-release-data}

\paragraph{Scores and source review.}
We screen for fifteen settings per column and then five observed scores per
row. We normalize bounded scores to $[0,1]$ using documented denominators,
infer fraction and percentage labels from source formatting, and exclude
unsupported scales. Multiple-choice scores retain their chance level.
During source review, we correct WritingBench and AIME denominators,
exclude language-specific and reasoning-budget results from pooled defaults,
and drop cells whose source spread exceeds 0.05 on the normalized scale.
The resulting \corpusname{} matrix contains \bobReleaseReviewedConfigs{}
configurations, \bobReleaseReviewedGroups{} release groups,
\bobReleaseBenchmarks{} benchmark columns and \bobReleaseReviewedCells{}
observations. The study archive links the intermediate snapshot and reviewed
matrix through the metadata and exclusions.

\paragraph{Release groups and benchmark families.}
We assign related configurations of a creator's base model to one release
group. Benchmark families are connected components of named versions and
documented derivations. MGSM links to GSM8K, MultiPL-E links HumanEval and
MBPP, and Multi-IF links to IFEval
\citep{shi2022mgsm,cassano2022multiple,he2024multiif}. Shared sources in
PolyMath and in visual suites such as MMStar connect larger families
\citep{wang2025polymath,chen2024mmstar,lu2023mathvista,zou2024dynamath,wang2024muirbench,liu2024ocrbench}.
We retain eligible versions as separate columns and use their
families for withholding and inference. The study archive records the
complete source graph and the frozen Qwen3-Embedding-8B descriptions,
paraphrases and vectors. Both BBH and BBEH pass the metadata screen. Their
lineage connection to the external BIG-bench data is described in
\Cref{sec:app-external}.

\paragraph{Splits and metrics.}
For seeds 20260916 and 20260917, we deterministically hash families into
five outer blocks each. To tune on the remaining columns, we use three
inner family blocks with the outer seed plus 101. We iteratively require
three observed cells per training row and column, then estimate geometry,
characteristic curves, residuals, information and clusters on that pool.
Outer-score perturbations leave these fits and selections unchanged.

For ranking, we require fifteen scored configurations and ten release
groups. We first average scores, predictions, and the training-percentile
reference $A_m$ within release groups (\Cref{sec:app-tests}). Level is the
Spearman correlation on these group means. Profile removes a quadratic in
the grouped reference from both ranks before correlation. For five-fold
isotonic target-score calibration, we hold out release groups,
require at least eight training groups and three distinct predictions,
and clip to the normalized range. Calibration metrics require ten test
groups and report errors on a 0--100 scale.

We match methods by split seed and target, then average seeds within
targets and targets within equally weighted families. We use 5,000 family
bootstrap draws for paired intervals and 100,000 draws for two-sided
sign-flip tests. The intervals describe variation across families for the
observed matrix and seeds. For the optional task comparison, we designate
ranking and regret at budget three as primary and apply Holm correction
to eleven continuous-minus-control contrasts per metric and fitting
schedule.

\subsection{Relevance controls and parameter selection}\label{sec:app-release-controls}

We compare relevance rules with shared curves, ability, residuals,
information, robust cell weights and $P/(P+1)$ regularization. The
continuous rule combines a Gaussian kernel with inverse density. At six
and forty-eight rounds, we tune each adjustable rule by maximizing
family-balanced inner Spearman on targets eligible under every candidate.
The support-matched top-$k$ rule rounds the continuous
rule's observed Kish effective-neighbor count and rescales the selected
neighbors to match its robust support mass per model. With zero local
support, the prediction is $\theta$.
The study archive provides the complete tuning grids and cluster settings.

\subsection{Retained-evidence controls}\label{sec:app-retained-controls}

We match evaluation coverage using saved six-round fits. The full fit
retains its larger training population. \Cref{sec:app-insights} also
matches the fitting rows. We adapt ten columns from the historical AA
v3.0 suite for the curated control, including text-only HLE and
LiveCodeBench v6 mappings
\citep{artificialanalysis2026methodology}. The control averages observed
standardized scores after family withholding. The published AA Index uses
its own evaluations and weighting rules.

On \bobMatchedTargets{} targets in \bobMatchedFamilies{} families, Spearman
is \bobMatchedThetaRank{} for global $\theta$ and \bobMatchedCuratedRank{}
for the curated control. The paired gap is \bobMatchedThetaCuratedDelta{}
[$\bobMatchedThetaCuratedLo{},\bobMatchedThetaCuratedHi{}$]. Full-pool
uniform averaging improved by \bobMatchedUniformCuratedDelta{} (Holm
$p=\bobMatchedUniformCuratedP{}$). PC1 and shrinkage-covariance GLS reach
\bobMatchedPCRank{} and \bobMatchedCovarianceRank{}, respectively.
We mean-impute standardized training scores for PC1. For GLS, we use
Ledoit--Wolf covariance with a 0.05 ridge and normalize weights over each
row's observations. The observed pairwise-correlation GLS covariance
remains non-positive-definite with that ridge. Shrinking row-centered,
mean-imputed residuals yields uniform weights because each residual row
sums to zero. These results motivate the direct-score covariance control.

The historical comparison has nine eligible ranking targets inside the
curated subset and 111 outside. We correct twelve ranking contrasts in
each of the three strata. Appendix Details provides matched
ranking, calibration and shortlist outcomes, including suite-specific
effects (Tables~9, 11 and~12 and Figure~6).

\begin{table}[H]\centering\small
\caption{Task support settings on \corpusname{}. The canonical row gives
the reference procedure for the main comparisons.}
\label{tab:release-settings}
\setlength{\tabcolsep}{4pt}
\begin{tabular}{llcccr}
\toprule
Support setting & Seeds & $h$ grid & $\lambda$ & $\alpha$ & $\rho$ \\
\midrule
Canonical Support & 2 & $H_5$ & 1 & 1 & 0.8232 \\
Fixed Support & 2 & $\{0.45\}$ & 1 & 1 & 0.8210 \\
Initial bandwidth search & 2 & $H_3$ & 1 & 1 & 0.8246 \\
Fixed, first seed & 1 & $\{0.45\}$ & 1 & 1 & 0.8248 \\
Initial joint search & 2 & $H_3$ & $L_4$ & $A_3$ & 0.8242 \\
Expanded joint search & 2 & $H_5$ & $L_5$ & $A_5$ & 0.8244 \\
\bottomrule
\end{tabular}

\par\smallskip\begin{minipage}{\linewidth}\footnotesize
All rows use six rounds, density multiplier 0.25, and the reviewed corpus.
$H_3=\{0.30,0.45,0.60\}$ and $H_5=\{0.15,0.225,0.30,0.45,0.60\}$ multiply
the training distance median.
$L_4=\{0.25,0.5,1,2\}$, $L_5=\{0.125,0.25,0.5,1,2\}$,
$A_3=\{0.25,1,4\}$ and $A_5=\{0.0625,0.125,0.25,1,4\}$.
Joint searches include $\theta$.
The canonical row selects $h$ by inner family-balanced ranking.
On common evaluation coverage (\Cref{sec:app-retained-controls}), the same
canonical predictions give \bobMatchedSupportRank{}.
\end{minipage}

\end{table}

\subsection{Retrospective shortlist design}\label{sec:app-release-decision}

In each outer fit, we select the configuration with the most observed
training scores in each release group, breaking ties by configuration ID.
For each target, we require at least ten representatives in the common
pool $C_q$ with observed outcomes and finite predictions under every
rule. We rank candidates by prediction and break ties by configuration
ID. At budgets
$k\in\{1,3,5\}$, hit means the nominated set $A_k(q)$ contains a candidate
tied for the best observed outcome. Regret is
$100[\max_{m\in C_q}s_{mq}-\max_{m\in A_k(q)}s_{mq}]$.
Both metrics use the observed candidate pool and assume evaluation of every
nominated candidate. The study archive provides candidate IDs, predictions and
scores.

\section{Evidence retention and benchmark multiplicity}
\label{sec:app-insights}

To compare evidence pools on identical fitting rows, we carry forward
the 354-by-148 matrix, fixed ten-column AA-v3-derived subset and outer
family blocks from \Cref{sec:app-retained-controls}.

\subsection{Populations, estimators and inference}

On training data alone, we recursively require three observed models
per column and remove constant columns. The strict panel requires
three curated observations per model. For the broad panel, we allow one
after the three-observation full-pool screen. Curated and full fits share
rows within each panel. The strict panel has 30--127 fitting models per
fold and 50 ranking targets in 29 families. The broad panel has 207--300
fitting models and 120 targets in 60 families. All forty observation
graphs are connected.

For each six-round fit, we recompute the median training distance and keep
the density multiplier at 0.25. Pool expansion thus changes available
scores and description geometry. Standardized averaging uses observed
cells. For PC1, we fill missing standardized values with zero and orient
the first component by observed row means.
The affine factor minimizes
$\sum_{(m,i)\in\Omega}[z_{mi}-(c_i+\ell_i t_m)]^2$ on observed training
scores standardized by column. The rank-one factor fixes $c_i=0$. Both
use alternating least squares, one of three deterministic starts selected
by reconstruction loss, and signed loadings. The source-family control
weights each column by the inverse size of its training family using
lineage metadata.

We require fifteen configurations and ten release groups for ranking and
use the paired family intervals and tests in \Cref{sec:app-release-data}.
We apply separate Holm corrections to seven factorial and six estimator
comparisons across both panels, both stopping schedules and three target
strata (84 and 72 tests, respectively).
Nine eligible ranking targets belong to the curated suite in each panel.
Outside the suite, density retention gains are
\bobStrictOutsideRetention{} (41 targets, 24 families, Holm
$p=\bobStrictOutsideHolm{}$) and \bobLowOutsideRetention{} (111 targets,
57 families, Holm $p=\bobLowOutsideHolm{}$).

\begin{table}[H]\centering\scriptsize
\caption{Six-round factorial contrasts on eligible targets. Holm $p$
covers all 84 factorial comparisons.}
\label{tab:insight-factorial}
\setlength{\tabcolsep}{4pt}
\begin{tabular}{llrrr}
\toprule
Suite support & Comparison & $\Delta\rho$ & 95\% CI & Holm $p$ \\
\midrule
$\geq3$ & Mean (full) $-$ Mean (curated) & +0.0076 & [-0.0185, +0.0365] & 1.0000 \\
$\geq3$ & Equal equating (full) $-$ Equal equating (curated) & +0.0367 & [+0.0208, +0.0529] & 0.0050 \\
$\geq3$ & Density equating (full) $-$ Density equating (curated) & +0.0350 & [+0.0183, +0.0520] & 0.0130 \\
$\geq3$ & Equal equating (curated) $-$ Mean (curated) & -0.0159 & [-0.0317, -0.0003] & 1.0000 \\
$\geq3$ & Density equating (curated) $-$ Equal equating (curated) & +0.0000 & [+0.0000, +0.0001] & 1.0000 \\
$\geq3$ & Equal equating (full) $-$ Mean (full) & +0.0132 & [-0.0074, +0.0337] & 1.0000 \\
$\geq3$ & Density equating (full) $-$ Equal equating (full) & -0.0016 & [-0.0102, +0.0076] & 1.0000 \\
$\geq1$ & Mean (full) $-$ Mean (curated) & +0.0226 & [-0.0035, +0.0492] & 1.0000 \\
$\geq1$ & Equal equating (full) $-$ Equal equating (curated) & +0.0342 & [+0.0180, +0.0496] & 0.0075 \\
$\geq1$ & Density equating (full) $-$ Density equating (curated) & +0.0364 & [+0.0186, +0.0540] & 0.0107 \\
$\geq1$ & Equal equating (curated) $-$ Mean (curated) & +0.0227 & [+0.0091, +0.0348] & 0.0598 \\
$\geq1$ & Density equating (curated) $-$ Equal equating (curated) & -0.0000 & [-0.0000, +0.0000] & 1.0000 \\
$\geq1$ & Equal equating (full) $-$ Mean (full) & +0.0343 & [+0.0122, +0.0566] & 0.2349 \\
$\geq1$ & Density equating (full) $-$ Equal equating (full) & +0.0022 & [-0.0023, +0.0067] & 1.0000 \\
\bottomrule
\end{tabular}

\end{table}

\begin{table}[H]\centering\scriptsize
\caption{Six-round estimator contrasts on eligible targets. Holm $p$
covers all 72 estimator comparisons.}
\label{tab:insight-estimators}
\setlength{\tabcolsep}{4pt}
\begin{tabular}{llrrr}
\toprule
Suite support & Comparison & $\Delta\rho$ & 95\% CI & Holm $p$ \\
\midrule
$\geq3$ & Equal equating (full) $-$ PC1 (full) & +0.0877 & [+0.0494, +0.1294] & 0.0043 \\
$\geq3$ & Equal equating (full) $-$ Masked factor (full) & +0.0098 & [-0.0062, +0.0271] & 1.0000 \\
$\geq3$ & Equal equating (full) $-$ Affine factor (full) & -0.0084 & [-0.0257, +0.0063] & 1.0000 \\
$\geq3$ & Masked factor (full) $-$ PC1 (full) & +0.0779 & [+0.0429, +0.1164] & 0.0071 \\
$\geq3$ & Affine factor (full) $-$ Masked factor (full) & +0.0182 & [+0.0044, +0.0324] & 0.6403 \\
$\geq3$ & Density equating (full) $-$ Family equating (full) & -0.0131 & [-0.0265, -0.0017] & 1.0000 \\
$\geq1$ & Equal equating (full) $-$ PC1 (full) & +0.1474 & [+0.0875, +0.2138] & 0.0007 \\
$\geq1$ & Equal equating (full) $-$ Masked factor (full) & +0.0205 & [+0.0040, +0.0366] & 0.6919 \\
$\geq1$ & Equal equating (full) $-$ Affine factor (full) & -0.0156 & [-0.0300, -0.0037] & 0.5072 \\
$\geq1$ & Masked factor (full) $-$ PC1 (full) & +0.1269 & [+0.0700, +0.1906] & 0.0014 \\
$\geq1$ & Affine factor (full) $-$ Masked factor (full) & +0.0361 & [+0.0200, +0.0549] & 0.0020 \\
$\geq1$ & Density equating (full) $-$ Family equating (full) & -0.0045 & [-0.0104, +0.0015] & 1.0000 \\
\bottomrule
\end{tabular}

\end{table}

\begin{table}[H]\centering\small
\caption{Full-minus-curated Spearman at 48 rounds on the same fitting
populations. The six-round contrasts are in \Cref{tab:insight-factorial}.}
\label{tab:insight-longer}
\begin{tabular}{@{}lrr@{}}
\toprule
Equating weights & Strict & Broad \\
\midrule
Equal & $+0.032$ & $+0.032$ \\
Density & $+0.032$ & $+0.033$ \\
Source-family & $+0.054$ & $+0.043$ \\
\bottomrule
\end{tabular}
\end{table}

Extending a capped standardized-affine fit to convergence leaves its
ranking metrics unchanged. Two raw-score affine fits are degenerate.
Appendix Details contains the full 48-round comparison (Table~15) and the numerical sensitivity results.

\subsection{Listing robustness and initialization}
\label{sec:app-insight-listings}

We add four copies of each of eight coverage-selected benchmarks, with
exact or paraphrased descriptions. Across the sixteen scenarios, we compare
standardized averaging, equal-weight equating, density and source-family
weights, and known-copy weights. A concurrent audit also shows that merging
substitute benchmarks can reorder an equal-weight ranking
\citep{navasardyan2026audit}.

\begin{table}[H]\centering\scriptsize
\caption{Rank displacement after four repeated listings. Means average
models and then eight scenarios. Worst is the largest single-model displacement.}
\label{tab:insight-listings}
\setlength{\tabcolsep}{4pt}
\begin{tabular}{lrrrr}
\toprule
& \multicolumn{2}{c}{Exact} & \multicolumn{2}{c}{Paraphrase} \\
Method & Mean & Worst & Mean & Worst \\
\midrule
Uniform & 5.907 & 127 & 5.907 & 127 \\
Equating, equal weights & 1.948 & 73 & 1.948 & 73 \\
Density, refitted bandwidth & 0.626 & 44 & 0.955 & 55 \\
Density, fixed bandwidth & 0.625 & 44 & 0.953 & 55 \\
Density, weighted initialization & 0.569 & 42 & 0.917 & 55 \\
Density, fixed bandwidth + weighted initialization & 0.567 & 42 & 0.912 & 55 \\
Source-family weights & 1.011 & 51 & 1.011 & 51 \\
Source-family weights + weighted initialization & 0.978 & 51 & 0.978 & 51 \\
Known-copy ID weights & 0.150 & 5 & 0.150 & 5 \\
Known-copy ID weights + weighted initialization & 0.000 & 0 & 0.000 & 0 \\
\bottomrule
\end{tabular}

\end{table}

For known copies, we split the original coefficient among the copies.
Using these weights in initialization and fitting yields zero rank
movement in all sixteen scenarios, with a maximum latent-score change of
$\bobListingIdentityMaxChange{}$. Soft density also changes neighboring
coefficients. At fixed bandwidth, adding $k$ copies to a benchmark of
density $\rho_q$ raises its group's unnormalized mass by
$k(\rho_q-1)/[\rho_q(\rho_q+k)]$. This quantity is zero for an isolated
benchmark (\Cref{sec:app-proof}). Under soft density, fixing bandwidth
and weighting initialization yield only modest reductions in movement.

\subsection{Independent measurements under fixed capability weights}
\label{sec:app-insight-simulation}

To isolate independent measurement error in weighted averages, we
simulate thirty model populations with repeated 80-item binomial scores.
We add exact copies, independent measurements of one of the original
eight capabilities, or measurements of a ninth capability. The weighted
score $\widehat T_m=\sum_d w_d\overline{s}_{md}$ uses fixed weights $w_d$
and $n_d$ measurements per capability. With independent errors within and
across capabilities, its variance is $\sum_d w_d^2\sigma_{md}^2/n_d$.

With known capability identities, four independent measurements reduce
MSE by 10.0\% at fixed capability weight $1/8$. Exact copies preserve
the original error. Under soft density, independent measurements reduce
error relative to copies. Changes in neighborhood mass can raise error
above the starting pool under the equal-capability target.
Appendix Details gives the full simulation (Table~17).

\section{Protocol comparability and partial reporting}
\label{sec:app-review}

\subsection{Metadata-restricted evidence}
\label{sec:app-review-metadata}

Before evaluating predictions, we fix a column-level metadata screen for
uncertain benchmark versions and settings, including generic LiveCodeBench
date windows, IFEval scoring levels, MMMLU language sets and SciCode
problem levels. The screen removes \bobMetadataExcluded{} columns and
\bobMetadataExcludedCells{} cells, leaving \bobMetadataRetained{} columns
and \bobMetadataRetainedCells{} cells. Prompt, generation-budget and harness
differences can remain. The curated suite retains AA-LCR, HLE text-only,
IFBench, MMLU-Pro and Terminal-Bench Hard.

We screen each fold before matching fitting rows across pools.
Three of twenty panel--seed--fold
combinations become ineligible. The strict rule leaves only
\bobMetadataStrictTargets{} targets in \bobMetadataStrictFamilies{}
families. The broad rule retains 52--199 fitting models in completed folds
and \bobMetadataLowTargets{} targets in \bobMetadataLowFamilies{} families.

\begin{table}[H]\centering\small
\caption{Retention after the metadata restriction. Full and curated fits
share model rows. Spearman averages 66 targets in 43 families under the
broad rule. Holm correction covers twenty retention comparisons.}
\label{tab:review-metadata}
\setlength{\tabcolsep}{4pt}
\begin{tabular}{lrrrrr}
\toprule
Estimation & Curated & Full & $\Delta\rho$ & 95\% CI & Holm $p$ \\
\midrule
Standardized mean & 0.754 & 0.760 & +0.006 & [-0.032, +0.040] & 1.000 \\
Equal equating & 0.791 & 0.813 & +0.022 & [-0.002, +0.050] & 1.000 \\
Density equating & 0.791 & 0.815 & +0.024 & [+0.002, +0.050] & 0.609 \\
Affine factor & 0.795 & 0.802 & +0.007 & [-0.018, +0.031] & 1.000 \\
Density affine & 0.795 & 0.804 & +0.009 & [-0.013, +0.031] & 1.000 \\
\bottomrule
\end{tabular}

\end{table}

On this restricted population, the density-equating gain is
\bobMetadataLowRetention{} and its corrected test is nonsignificant
(Holm $p=\bobMetadataLowHolm{}$).
\Cref{sec:app-semantic-families} separates training restriction from
evaluation-target exclusion.

\subsection{Partial copies and estimator transfer}
\label{sec:app-review-partial}

For each of eight copy targets, we append four copies with a shared mask.
We retain 25\%, 50\% or 75\% of models reporting the original score under
three seeds, or use the benchmark mask closest to 50\% overlap by
missingness alone. With complete copies, the design yields 88 scenarios.
We keep the original bandwidth fixed.

Column density reduces mean movement at 50\% and 75\% coverage. At 25\%,
mean movement is \bobPartialQuarterDensity{} under density and
\bobPartialQuarterFlat{} under equal-weight equating (\cref{tab:partial-main}
and \emph{Appendix Details}, original Table 19). Among models without
copies at that coverage, within-group ordering changes by
\bobPartialQuarterUnexposedWithin{} places and original-cell fitted
predictions by \bobPartialQuarterUnexposedFit{} score points on average.
Observed mass depends on reporting even for an isolated benchmark.
Assigning $1/(k+1)$ to each of $k+1$ listings gives a model with $r_m$
copies mass $(1+r_m)/(k+1)$. With four copies, a model reporting only
the original gets one fifth of the mass of a model reporting all five.
Partial columns also lead to curve fits on different rows.

\begin{table}[H]\centering\small
\caption{Mean absolute rank displacement under affine fitting on 127
supported models. Both rules use the same 84 finite scenarios and fixed
bandwidth.}
\label{tab:review-supported}
\setlength{\tabcolsep}{4pt}
\begin{tabular}{lrrr}
\toprule
Copy coverage & Cases & Affine & Density affine \\
\midrule
Full & 8 & 1.041 & 0.220 \\
25\% & 23 & 0.546 & 0.567 \\
50\% & 22 & 0.709 & 0.437 \\
75\% & 24 & 0.905 & 0.289 \\
Reporting pattern & 7 & 1.030 & 0.531 \\
\bottomrule
\end{tabular}

\end{table}

For the density-affine fit, we weight the observed-cell loss by column
density. Both affine rules have degenerate loading updates on the full
matrix. On 127 supported models and 131 columns, three masks yield
constant copied scores and one equal-affine fit fails. In the other 84
cases, density weighting reduces full-copy movement from
\bobSupportedFullAffine{} to \bobSupportedFullDensityAffine{} places.
At 25\% coverage, movement is
\bobSupportedQuarterDensityAffine{} with density and
\bobSupportedQuarterAffine{} with equal weights. The study archive provides
all failed fits, worst cases and bandwidth checks.

\section{Global grouping and observed support distributions}
\label{sec:app-semantic}

\subsection{Effective-mass-matched semantic grouping}
\label{sec:app-semantic-grouping}

We compare density weighting and complete-linkage grouping on the same
description distances. In each fitting pool, density weighting uses
$h=0.25\,\operatorname{median}_{i<j}d_{ij}$ and effective mass
$W=\sum_i\rho_i^{-1}$. To match the rounded mass, we cut the hierarchy
at $G=\min(N,\max(1,\lfloor W+1/2\rfloor))$ clusters and assign column
$i$ in cluster $c(i)$ weight $u_i=1/|c(i)|$ before normalization. This
match allows neighborhood sizes and weight distributions to differ.
On 148 columns, $W=\bobSemanticMass{}$ gives \bobSemanticGroups{}
clusters. A cut at the kernel half-height would give
\bobSemanticHalfGroups{}.

For both rules, we refit curves and robust weights on the same rows for
six rounds, with ordinary or weighted initialization. The full-pool
marginal interval favors density weighting in the strict panel and spans
zero in the broad panel. Curated-pool differences are small. With
unweighted initialization, grouping's full-pool Spearman gains are 0.0257
and 0.0328 in the strict and broad panels.
Appendix Details gives the full contrasts and family bootstrap intervals
(Tables~21--22).

We use 96 complete, partial or paraphrased copy scenarios at fixed
bandwidth, recomputing clusters from the expanded descriptions.
Mean rank displacement favors density in each
condition. Worst-case comparisons are mixed. At 25\% coverage, both
rules move ranks more than equal-weight equating
(\Cref{tab:partial-main}). In a fixed cluster with $n$ measurements,
$k$ copies increase the original measurement's share from $1/n$ to
$(k+1)/(n+k)$.

\begin{table}[H]\centering\scriptsize
\caption{Mean absolute rank displacement for the two column weighting
rules. Parentheses give the largest displacement.}
\label{tab:semantic-listing}
\setlength{\tabcolsep}{4pt}
\begin{tabular}{lrrrrr}
\toprule
& & \multicolumn{2}{c}{Unweighted initialization} & \multicolumn{2}{c}{Weighted initialization} \\
Condition & Cases & Density & Grouping & Density & Grouping \\
\midrule
Complete exact & 8 & 0.625 (44) & 0.876 (45) & 0.567 (42) & 0.809 (41) \\
Complete paraphrase & 8 & 0.953 (55) & 1.440 (94) & 0.912 (55) & 1.406 (90) \\
25\% coverage & 24 & 0.761 (42) & 0.799 (47) & 0.758 (41) & 0.787 (45) \\
50\% coverage & 24 & 0.706 (51) & 0.810 (51) & 0.697 (49) & 0.778 (41) \\
75\% coverage & 24 & 0.657 (52) & 0.862 (51) & 0.615 (50) & 0.810 (46) \\
Reporting pattern & 8 & 0.727 (50) & 0.869 (52) & 0.706 (50) & 0.866 (49) \\
\bottomrule
\end{tabular}

\end{table}

\subsection{Reporting overlap among semantic neighbors}
\label{sec:app-semantic-overlap}

\paragraph{A constraint on column-only weighting.}\label{sec:app-mask-constraint}
Consider two models that report the same scores, with one model also
reporting a new listing of a benchmark of weight $v>0$. Their other
benchmarks have common total weight $B$ before and $B'$ after adding the
listing. With positive original and new weights $a$ and $b$, preserving
the benchmark's relative weight for both models would require
\[
 a/B'=v/B,\qquad (a+b)/B'=v/B .
\]
The equations imply $b=0$. Model-specific weights allow different
coefficients for the two models. The constraint concerns nominal weights,
and rankings also depend on curves, robust weights and refitting.
See \citet{paruolo2013ratings,nardo2008handbook} for related distinctions
between assigned weights and realized importance in composite indices,
and \citet{sanderson2017skill} for inverse-similarity weights in climate
ensembles.

We measure overlap on individual configurations. With $O_i$ the set of
models reporting benchmark $i$, directional coverage is
$C(j\mid i)=|O_i\cap O_j|/|O_i|$ and Jaccard overlap is
$J(i,j)=|O_i\cap O_j|/|O_i\cup O_j|$. We use one nearest neighbor per
column and, in a separate distribution, weight all directed pairs by
kernel similarity.

\begin{table}[H]\centering\small
\caption{Reporting overlap among description neighbors. The
$C\leq25\%$ column gives the fraction of directions or kernel mass.}
\label{tab:semantic-overlap}
\setlength{\tabcolsep}{4pt}
\begin{tabular}{lrrrrr}
\toprule
Pair distribution & Directions & Mean $C$ & Median $C$ & $C\leq25\%$ & Median $J$ \\
\midrule
Nearest neighbor & 148 & 39.2 & 36.1 & 44.6 & 20.1 \\
Kernel-weighted pairs & 21,756 & 33.5 & 20.3 & 56.9 & 10.0 \\
$K\geq0.5$ pairs & 16 & 37.4 & 22.2 & 62.5 & 10.2 \\
\bottomrule
\end{tabular}

\end{table}

Of 148 nearest-neighbor directions, \bobOverlapNearestLow{} have
coverage at most 25\%. The $K\geq0.5$ set contains
\bobOverlapStrongPairs{} unordered pairs across
\bobOverlapStrongBenchmarks{} benchmarks. Appendix Details plots both
distributions (Figure~7). These overlaps locate
the configurations that share observations. Estimating how often
natural re-listings change rankings also requires their scores,
settings and reporting patterns.

\subsection{Metadata-restricted gains across evaluation families}
\label{sec:app-semantic-families}

The broad panel starts with 120 targets in 60 families and retains 93 in
53 families on the restricted fitting rows, then 66 in 43 families after
target exclusion. On these final rows and targets, both point estimates
rise when we restrict the training columns. The larger gain for the
curated pool narrows the retention gap, reflecting combined changes to
evidence and suite support.

\begin{table}[H]\centering\small
\caption{Density-equating Spearman on identical restricted fitting
rows and the same 66 targets in 43 families.}
\label{tab:semantic-metadata-cells}
\setlength{\tabcolsep}{4pt}
\begin{tabular}{lrrr}
\toprule
Training pools & Curated & Full & Full $-$ curated \\
\midrule
Original & 0.7433 & 0.7941 & +0.0508 \\
Restricted & 0.7908 & 0.8149 & +0.0241 \\
\bottomrule
\end{tabular}

\end{table}

The restricted retention estimate is \bobMetadataLowRetention{} (Holm
$p=\bobMetadataLowHolm{}$). Of the 43 family effects,
\bobFamilyPositive{} are positive and \bobFamilyNegative{} negative, with
median \bobFamilyMedian{}. Leave-one-family-out estimates with fixed
predictions range from \bobFamilyDeleteLo{} to \bobFamilyDeleteHi{}.
Appendix Details gives the complete support counts and family
distribution (Table~25 and Figure~8).

\subsection{Shared task design with distinct length-specific measurements}
\label{sec:app-ruler-case}

RULER shares retrieval, tracing, aggregation and question-answering
designs across six context lengths. For its question-answering tasks,
distractors lengthen contexts from SQuAD and HotpotQA
\citep{hsieh2024ruler}. Each length has a released score, and all six pass
the metadata screen. Under the original full-pool geometry, the mean-one
coefficients $n u_i/W$ range from \bobRulerMinCoefficient{} to
\bobRulerMaxCoefficient{} and total \bobRulerGroupCoefficient{}, compared
with six under equal weighting. These lengths occupy six of the seven
lowest coefficient positions. The source scores in \Cref{tab:ruler-case}
illustrate the distinct measurements that share this influence.

\begin{table}[H]\centering\small
\caption{RULER reporting coverage, mean-one coefficients and
Apertus-8B-Instruct scores from Table~23 of
\citet{projectapertus2025apertus}. A dash denotes an unreported score.}
\label{tab:ruler-case}
\begin{tabular}{@{}lrrr@{}}
\toprule
RULER context & Observed models & Mean-one coefficient & Example score (\%) \\
\midrule
4K & 22 & 0.479 & 91.2 \\
8K & 22 & 0.497 & 87.2 \\
16K & 22 & 0.402 & 79.1 \\
32K & 26 & 0.426 & 65.9 \\
64K & 25 & 0.388 & 61.4 \\
128K & 29 & 0.389 & -- \\
\bottomrule
\end{tabular}

\end{table}

To measure whether the other lengths help predict a held-out RULER
ranking, we exclude the family from the anchor fit and add the other
lengths to the related fit. We hold all 354 fitting rows, the original
bandwidth and the six-round schedule fixed, then compare global $\theta$
with held-out score ranks. In the full pool, Spearman agreement between
BoB and equal-weight rankings is \bobRulerRankAgreement{} among the 32
configurations with any RULER score and 0.9982 among the 19 with every
length. Adding related lengths improves held-out agreement under both
rules, with a larger mean gain under equal weighting. The comparison
on BBH and its harder successor BBEH yields mixed differences between
methods, and five configurations report both \citep{kazemi2025bbeh}.
\emph{Appendix Details} gives the per-target fits (original Table~28).

\begin{table}[H]\centering\small
\caption{Global $\theta$ correlations with held-out related benchmarks.
The gain row subtracts the family-excluded anchor.}
\label{tab:natural-signal}
\begin{tabular}{@{}lrr@{}}
\toprule
Held-out target or contrast & BoB & Equal equating \\
\midrule
RULER, mean across six lengths & \bobRulerHeldoutDensity{} & \bobRulerHeldoutEqual{} \\
RULER, mean gain from other lengths & \bobRulerRelatedDensity{} & \bobRulerRelatedEqual{} \\
BBH & 0.8670 & 0.8449 \\
BBEH & 0.9643 & 0.9679 \\
\bottomrule
\end{tabular}
\end{table}

\section{Optional task-conditioned comparison}
\label{sec:app-task-extension}

A task description can guide model selection.
Learned model and task representations support score prediction and routing
\citep{zhuang2024embedllm,song2025irtrouter,chen2025irtnet,ge2026agentpsych},
and text-only forecasts use task descriptions and model configurations
\citep{park2025anticipatory}. Taxonomies and nearest neighbors define discrete
relations among tasks \citep{li2025knnrouter}. We adjust global ability with
smoothly varying benchmark relevance.

\subsection{Task-conditioned residual aggregation}\label{sec:field}

We embed the query description $q$ in the shared space and let
$g_{mi}(q)\geq0$ denote benchmark relevance. The prediction $y_m^{(g)}(q)$
adds a regularized mean residual to model $m$'s fitted ability.
\begin{equation}\label{eq:field}
 \begin{aligned}
 \bar r_m^{(g)}(q)&=\frac{\sum_i g_{mi}(q)\bar I_iw_{mi}r_{mi}}
                            {\sum_i g_{mi}(q)\bar I_iw_{mi}},\qquad
 P_m^{(g)}(q)=\sum_i g_{mi}(q)w_{mi},\\
 y_m^{(g)}(q)&=\theta_m+\beta_m(q)\bar r_m^{(g)}(q).
 \end{aligned}
\end{equation}
The residual mean uses curve information $\bar I_i$ and robust cell weights
$w_{mi}$, with zero weight for unobserved cells. Support $P_m^{(g)}(q)$ measures
the amount of relevant evidence reported. We set
continuous relevance to $g_{mi}(q)=\exp[-d(q,x_i)^2/(2h^2)]u_i$, so distance
attenuates relevance and co-located entries share column mass. \bobsupport{} uses
$\beta_m=\lambda P_m^{(g)}/(P_m^{(g)}+\alpha)$, with $\lambda=\alpha=1$ in the
unit-scale rule. \bobconstant{} fixes $\beta_m=c$. Both return $\theta_m$ at
zero support. To isolate neighborhood selection from correction strength,
we use hard relevance and support-matched top-$k$ controls
(\Cref{sec:app-relevance-controls,sec:app-aggregation}).

\subsection{Queries and score dependence}\label{sec:exogeneity}

We withhold target scores in each outer fit and select tunable parameters in
inner folds for \corpusname{} and the mechanism-matched AA comparisons.
For the original AA reference, we select relevance bandwidth using outer
outcomes. Fully nested AA selection reaches the same reported Level and Profile
(\Cref{sec:app-aggregation,sec:app-sensitivity}). Density coefficients depend on
the retained descriptions and density scale. Relevance-bandwidth selection
also uses training outcomes. \Cref{sec:app-canon} describes and evaluates
query canonicalization.

\subsection{Ranking and shortlist results}\label{sec:release-shortlists}

On the original \corpusname{} population, the paired gain in held-out ranking
over the matched global fit at \bobReleaseThetaRank{} is
\bobReleaseReviewedThetaRankDelta{}
[\bobReleaseReviewedThetaRankLo{}, \bobReleaseReviewedThetaRankHi{}].
Within the main grid, \bobconstant{} exceeds unit-scale \bobsupport{}
(Holm $p=\bobReleaseReviewedConstantRankHolm{}$), and the support-matched
top-$k$ contrast is nonsignificant (Holm
$p=\bobReleaseReviewedMatchedRankHolm{}$).

\begin{table}[H]
\centering\small
\caption{Optional task correction on the original \corpusname{} population.
Ranking covers \bobReleaseRankTargets{} targets in \bobReleaseRankFamilies{} families.
Paired $\Delta\rho$ uses global $\theta$ as reference. MAE and Regret@3 (R@3) use
0--100 score points. Hit@3 (H@3) is a percentage. Eligibility is shared within each
metric. MAE calibration withholds the evaluated release group.}
\label{tab:release-main}
\setlength{\tabcolsep}{4pt}
\begin{tabular}{lrrrrr}
\toprule
Method & $\rho$ & $\Delta\rho$ vs. $\theta$ & MAE & R@3 & H@3 (\%) \\
\midrule
Uniform standardized averaging & 0.764 & -0.053 & 6.19 & 2.08 & 65.3 \\
BoB global ($\theta$ only) & 0.817 & +0.000 & 5.18 & 1.62 & 71.7 \\
\bobsupport{} & 0.823 & +0.006 & 5.10 & 1.67 & 72.6 \\
\bobconstant{} & 0.830 & +0.014 & 5.04 & 1.52 & 74.4 \\
\bottomrule
\end{tabular}

\end{table}

For shortlist evaluation, we select release-group representatives by training
coverage. Regret is the gap to the best observed score. Hit records whether
the shortlist includes a best candidate. Regret@3 falls
from \bobReleaseUniformRegretThree{} under uniform standardized averaging to
\bobReleaseBobRegretThree{} under \bobsupport{}, with a paired reduction after
multiplicity correction. The paired change from global \bobname{} at
\bobReleaseThetaRegretThree{} is \bobReleaseReviewedThetaRegretDelta{} (Holm
$p=\bobReleaseReviewedThetaRegretHolm{}$). The gain in ranking thus accompanies
similar shortlist regret. We calibrate absolute scores using target
observations from other release groups. The task and shortlist results cover
benchmark-like descriptions and observed targets, with prospective cost and
latency outside the evaluation. \Cref{sec:app-release-results} reports numerical
sensitivity and indexes the full shortlist comparisons.

\section{\corpusname{} task correction and sensitivity}\label{sec:app-release-results}

Ranking improves over the matched global control with task correction.
Constant coefficients remain competitive across fitting rounds and local
scales. Absolute-score calibration, ranking, and shortlists use separate
eligibility rules. A target can support a shortlist even when its scores or
configuration count yield an undefined rank correlation. Appendix Details,
\S\,G gives all mechanisms, shortlist
budgets, domains and paired estimates (original Tables 30--38 and Figure 9).
For the primary six-round comparisons, we average effects by family and apply
Holm correction over eleven contrasts per metric. For sensitivity analyses,
we use unadjusted paired family-bootstrap intervals.

\subsection{Sensitivity and numerical stability}\label{sec:app-release-stability}

For the six-round fit (\Cref{tab:release-settings}), we select local bandwidth
on inner folds. Joint selection of bandwidth, strength, and support scale
in the initial nested grid improves mean held-out ranking by $+0.00745$ Spearman
[$+0.00349$, $+0.01220$] over global $\theta$. The expanded grid favors the
constant rule over the joint rule by $0.00593$ [$0.00198$, $0.01000$]. The
48-round joint comparison with repeated selection improves on its global
control by $+0.00599$ [$+0.00184$, $+0.01092$], and joint selection over 6,
12, 24 and 48 rounds improves on stopping-tuned $\theta$ by $+0.00885$
[$+0.00471$, $+0.01396$]. The initial results prompted us to extend these grids
and the refinement and stopping checks. All dependent analyses use the
corrected source blocks and tied initialization ranks.

At fixed local scales, density changes ranking and calibration by different
amounts. The metrics are identical across robust-weight floors tested.
We add four copies to each of eight source benchmarks with identical or
paraphrased descriptions. Mean absolute displacement is lower than under
uniform standardized averaging across density scales tested.
As fitting continues through 192 rounds, parameter
updates shrink and fixed-rule held-out prediction declines. At 48 rounds,
refining the grid step from 0.01 to 0.005 changes ranking by
$-0.00133$ [$-0.00240$, $-0.00043$]. The metrics change little with wider
ability limits. Some curve parameters still reach their bounds.
Appendix Details, \S\,G gives all tested grids and the study archive
records the per-round parameter changes and optimizer status.

\section{Artificial Analysis data and controls}\label{sec:app-aa}

We predict target scores with the optional \bobsupport{} correction
(\Cref{sec:app-task-extension}). Appendix Details, \S\,H gives the full
specifications, tables, and illustrations.

\subsection{Data and baselines}\label{sec:app-corpora}

We retrieved the Artificial Analysis Data API snapshot
(\url{https://artificialanalysis.ai/api/v2/data/llms/models}) at
\texttt{\bobAASnapshotRetrieved{}}. The JSON has \bobAASnapshotRows{} rows
and SHA-256 digest \texttt{\bobAASnapshotSha{}}. The fourteen fields are AIME,
AIME'25, MATH-500, GPQA Diamond, HLE, MMLU-Pro, LiveCodeBench, SciCode, IFBench,
long-context reasoning, Terminal-Bench Hard and 2.1, $\tau^2$-Bench, and
$\tau$-Bench Banking. Each row denotes one reported model configuration.
We treat nulls as missing and retain numeric values, including zero.
Requiring at least five numeric fields leaves \bobAARetainedRows{} rows.
Coverage by benchmark is in Appendix Details, \S\,H (original Table 39).

\begin{table}[H]
\centering\small
\caption{Original AA target-prediction reference. Means across fourteen
folds are shown with standard errors. The original bandwidth uses fold
outcomes for selection. Fully nested selection reaches the same Level and
Profile. The published AA Index is a descriptive comparator with its
original inputs.}
\label{tab:headline}
\setlength{\tabcolsep}{4pt}
\begin{tabular}{lcc}
\toprule
Prediction method & Level $\uparrow$ & Profile $\uparrow$ \\
\midrule
\bobsupport{} (ours) & \textbf{0.904}\,{\small (0.014)} & \textbf{0.483}\,{\small (0.051)} \\
Uniform standardized averaging & 0.878\,{\small (0.016)} & 0.046\,{\small (0.042)} \\
Raw-score category mean & 0.864\,{\small (0.018)} & 0.303\,{\small (0.065)} \\
Raw-score nearest benchmark & 0.762\,{\small (0.027)} & 0.111\,{\small (0.037)} \\
AA Intelligence Index & 0.883\,{\small (0.019)} & 0.351\,{\small (0.095)} \\
\addlinespace
\multicolumn{3}{l}{\emph{Methods fitted from the current score population}} \\
Score-covariance weights & 0.697\,{\small (0.058)} & 0.123\,{\small (0.068)} \\
First principal component & 0.823\,{\small (0.022)} & 0.026\,{\small (0.083)} \\
\bottomrule
\end{tabular}

\end{table}

\subsubsection{Baseline specification}\label{sec:app-baselines}

We standardize pool scores for uniform averaging. Category averaging uses
six study-defined labels, with uniform averaging for singleton categories.
Nearest-benchmark prediction uses a model's closest observed pool score.
We fit score-covariance GLS with $w\propto\Sigma^{-1}\mathbf{1}$ and a 0.05
ridge on row-centered standardized pool scores. For PC1, we impute column
means and align the sign with the mean score. The frozen API snapshot
does not record the Index version. We use the ten historical v3.0 columns
as a descriptive split of the supplied Index \citep{artificialanalysis2026methodology}.
On 248 complete rows, their raw mean agrees with the Index at Spearman 0.938.

\subsubsection{Mechanism-matched aggregation controls}\label{sec:app-aggregation}

To isolate relevance $g$ in \Cref{eq:field}, we hold curves, ability,
residuals, mean item information, robust weights, regularization, and
missing-cell treatment fixed. We define hard neighborhoods by category,
nearest benchmark, top-$k$, radius, or inferred clusters. For support-matched
top-$k$, we take the rounded Kish effective count of the continuous weights
and rescale equally weighted neighbors to match the total support mass.
We select parameters by mean inner-fold Profile, scaling by the median
distance among each outer target's thirteen pool benchmarks.
The selected bandwidth for continuous relevance is 0.45 times that median
in all fourteen folds.

Profile reaches 0.483 with continuous relevance and 0.454 with support-matched
top-$k$ (Holm $p=0.104$). Category residuals return global ability in two
singleton folds. Excluding them leaves a mean continuous-minus-category
Profile difference of \bobNonSingletonProfileThree{} across twelve targets
(one-sided Wilcoxon $p=\bobNonSingletonProfilePThree{}$). Appendix Details,
\S\,H gives the controls, paired effects and the 19-row addition that
preserves all 5,327 shared observations (original Tables 41, 47 and 48).

\subsection{Ablations and population shift}\label{sec:app-aa-additional}
\subsubsection{Matched and standalone controls}\label{sec:experiments-ablation}

\begin{table}[H]
\centering\small
\caption{AA constructions and matched controls. Rows A--D compare complete
procedures with construction-specific ability controls. Rows below
\bobsupport{} share its evaluation path. Ability shift is the mean
scenario-wise maximum absolute shift in standardized ability after four copies.}
\label{tab:ablation}
\setlength{\tabcolsep}{4pt}
\begin{tabular}{lcccccc}
\toprule
Construction / change & Curves & Field & Shrink. & Level $\uparrow$ & Profile $\uparrow$ & Ability shift $\downarrow$ \\
\midrule
A: Linear $z$, precision mean & $-$ & $-$ & $-$ & 0.891 & 0.041 & 0.754 \\
B: Characteristic curves & $+$ & $-$ & $-$ & 0.898 & 0.194 & 0.682 \\
C: Linear $z$ field & $-$ & $+$ & $-$ & 0.896 & 0.100 & 0.337 \\
D: Curves + residual field & $+$ & $+$ & $-$ & 0.837 & 0.296 & 1.912 \\
\midrule
\bobsupport{} & $+$ & $+$ & $+$ & \textbf{0.904} & \textbf{0.483} & \textbf{0.095} \\
\midrule
$-$ shrinkage & $+$ & $+$ & $-$ & 0.845 & 0.392 & -- \\
\bobconstant{} (factor $\to c$) & $+$ & $+$ & const. & 0.907 & 0.483 & -- \\
$-$ density weighting ($u_i\equiv1$) & $+$ & $+$ & $+$ & 0.899 & 0.473 & 0.963 \\
\bottomrule
\end{tabular}

\end{table}

For the AA constant, we fix the coefficient to each fold's median support
factor. Its reported Profile equals that of \bobsupport{}. We select the
\corpusname{} constant in inner folds.
Profile is \bobFlatProfileThree{} with flat weighting and
\bobHeadlineProfileThree{} with density weighting
(one-sided $p=\bobFlatProfilePThree{}$). Ability shift rises from
\bobCopyShiftThree{} to \bobFlatCopyShiftThree{} standard deviations under
flat weighting. To test semantic placement, we shuffle descriptions
across 200 full refits at the selected bandwidth. Shuffled Profile averages
\bobShuffleMeanThree{} and peaks at \bobShuffleMaxThree{}, below the observed
value in each shuffle.

\subsubsection{Global-ranking stability}\label{sec:experiments-invariance}

We group rows by the API's \texttt{model\_creator.name} and add
\bobPopulationMinAdded{}--\bobPopulationMaxAdded{} configurations from one
developer. We refit score-side quantities with fixed descriptions. Spearman
agreement on the original models' rankings is at least 0.9999 with
\bobname{} and 0.9930--0.9986 with score-covariance weights (Appendix Details,
\S\,H, original Table 43).

\subsection{Held-out evaluation}\label{sec:app-tests}

Level measures Spearman agreement between predicted and observed target
rankings. For Profile, we use the unweighted mean $A_m$ of model $m$'s
training-benchmark midpoint percentiles over observed entries, with average
ranks for ties. Among the target's evaluable models, we regress predicted
and observed target ranks separately on $(1,A_m,A_m^2)$ and correlate the
residuals with Spearman. Profile is relative to $A_m$, which varies with the
training pool and can contain domain-specific strengths. An isotonic control
allows broader monotone dependence. With isotonic
residualization, $\theta$ reaches \bobThetaIsotonicThree{} and the correction
reaches \bobStrictIsotonicThree{} ($p=\bobThetaIsotonicPFour{}$).
For the primary 10-fold sensitivity analysis, we exclude AIME/AIME'25 and
Terminal-Bench Hard/2.1. Contrasts against \bobClosePairHolmSurvivors{} survive
Holm correction.
On the four targets outside the historical v3 suite, \bobsupport{} reaches
\bobOutIndexLevelThree{} Level and \bobOutIndexProfileThree{} Profile,
versus 0.887 and \bobOutIndexAAProfileThree{} for the supplied AA Index.
This comparison describes historical membership. Appendix Details, \S\,H
reports full paired tests and alternative adjustments (original Tables 44--46).

\subsection{Residual and support scales}\label{sec:app-ablation}
\label{sec:app-validity}

For a latent-scale alternative, we divide each cell residual by its curve
derivative and weight by local information. Level and Profile reach
\bobLatentLevelThree{} and \bobLatentProfileThree{}, respectively. Adding
curve information to the support factor yields \bobInfoLevelThree{} and
\bobInfoProfileThree{} (paired $p=\bobInfoProfilePTwo{}$ for Profile).

\subsection{Text and representation robustness}\label{sec:app-lineage}

With independent paraphrases, Level is \bobParaLevelThree{} and Profile is
\bobParaProfileThree{}. When we remove cross-benchmark references from four
descriptions, AIME/AIME'25 remains closest and the Terminal-Bench variants
second, with Level \bobStrippedLevelThree{} and Profile \bobStrippedProfileThree{}.
The $\tau$ pair is farther apart at \bobTauPairDistThree{}
(\bobTauPairStrippedDistThree{} after stripping), near the median
\bobMedianPairDistTwo{}, despite shared tool-use and user-simulation protocols.

\subsubsection{Embedding size}\label{sec:app-robustness}

Across the 0.6B, 4B and 8B Qwen3 embeddings \citep{qwen3embedding2025}, Level
and Profile vary by at most \bobEmbeddingHeadlineSpreadBoundThree{}.
Four-copy maximum rank displacement is \bobEmbeddingSmallCopyRankOne{},
\bobEmbeddingMediumCopyRankOne{}, and \bobEmbeddingLargeCopyRankOne{} ranks,
respectively, versus 98.9 under equal weighting (Appendix Details, \S\,H,
original Table 49).

\subsection{Bandwidth and numerical sensitivity}\label{sec:app-sensitivity}

The fully nested control yields 0.904 Level and \bobAggregationProfileThree{}
Profile using the selected 0.45 training-median relevance scale. The original
outer-outcome selection uses a different held-out criterion and gives the same
reported means. For density, the 0.25-median rule gives bandwidth 0.224 and
effective mass 12.5, compared with 14.0 at 0.05 times the median and
\bobDensityMassAtMedianOne{} at the median (Appendix Details, \S\,H,
original Figure 11). Robust cap and floor sweeps move Level and Profile by at
most \bobRobustLevelBoundThree{} and \bobRobustProfileBoundThree{}.
Residual standard deviations are floored at $10^{-4}$ and information uses
$\max(\sigma_i^2,10^{-6})$ in the denominator.

\subsection{Query canonicalization}\label{sec:app-canon}

Before embedding, we rewrite queries in the benchmark-description register
of subject, format, and difficulty. Four example queries with matching families
move closer to the benchmark list, and the poetry query remains farthest away.
The support coefficient follows that proximity and determines correction
strength. Appendix Details, \S\,H gives the query data (original Table 50).

\subsection{Illustrative software-agent query}
Appendix Details, \S\,H gives the query, its canonicalized description and its
task-conditioned ranking.

\subsection{Global-ranking comparison}\label{sec:app-top-rankings}

Across the 603 model configurations with both scores, the two full rankings agree at Spearman $\rho=0.970$ and Kendall $\tau=0.862$. The \emph{Observed} column counts non-missing scores among the fourteen analyzed benchmarks.

\begin{table}[H]
\centering
\scriptsize
\setlength{\tabcolsep}{2.5pt}
\caption{Union of the top 20 model configurations under the global \bobname{} ability estimate $\theta_m$ and the AA Intelligence Index. Bold indicates top-20 membership. This is a descriptive comparison of complete rankings, not a target-withheld prediction test.}
\label{tab:top-rankings}
\begin{tabular*}{\linewidth}{@{\extracolsep{\fill}}p{0.50\linewidth}rrrrr@{}}
\toprule
Model configuration & \shortstack{Benchmarks\\observed} & \bobname{} rank & $\theta_m$ & AA rank & AA Index \\
\midrule
Claude Fable 5.1 (Adaptive Reasoning, Max Effort, Default Fallback) & 6 & \textbf{1} & 1.778 & \textbf{1} & 65.7 \\
Claude Fable 5.1 (Adaptive Reasoning, Xhigh Effort, Default Fallback) & 6 & \textbf{2} & 1.739 & \textbf{2} & 64.8 \\
Muse Spark 1.3 (max) & 6 & \textbf{3} & 1.685 & \textbf{6} & 62.1 \\
Claude Fable 5.1 (Adaptive Reasoning, High Effort, Default Fallback) & 6 & \textbf{4} & 1.659 & \textbf{4} & 62.5 \\
Claude Opus 5 (Adaptive Reasoning, Xhigh Effort) & 6 & \textbf{5} & 1.630 & \textbf{4} & 62.5 \\
Claude Opus 5 (Adaptive Reasoning, Max Effort) & 6 & \textbf{6} & 1.627 & \textbf{3} & 63.1 \\
Claude Opus 5 (Adaptive Reasoning, High Effort) & 6 & \textbf{7} & 1.625 & \textbf{8} & 61.5 \\
Muse Spark 1.3 (xhigh) & 6 & \textbf{8} & 1.623 & \textbf{13} & 60.8 \\
Kimi K3 (max) & 6 & \textbf{9} & 1.620 & \textbf{17} & 59.7 \\
Gemini 3.8 Flash (high) & 6 & \textbf{10} & 1.609 & 22 & 58.7 \\
GPT-6 Astra (xhigh) & 6 & \textbf{11} & 1.609 & \textbf{10} & 61.0 \\
GPT-6 Astra (max) & 6 & \textbf{12} & 1.608 & \textbf{9} & 61.2 \\
GPT-5.6 Sol (max) & 9 & \textbf{13} & 1.607 & \textbf{11} & 60.9 \\
Claude Fable 5.1 (Adaptive Reasoning, Medium Effort, Default Fallback) & 6 & \textbf{14} & 1.605 & \textbf{14} & 60.5 \\
Grok 4.6 (high) & 6 & \textbf{15} & 1.596 & \textbf{11} & 60.9 \\
GPT-6 Astra (high) & 6 & \textbf{16} & 1.587 & \textbf{15} & 60.3 \\
Claude Fable 5 (Adaptive Reasoning, Max Effort, Opus 4.8 Fallback) & 9 & \textbf{17} & 1.570 & \textbf{6} & 62.1 \\
GLM-5.3 (max) & 6 & \textbf{18} & 1.570 & \textbf{18} & 59.5 \\
Gemini 3.8 Flash (medium) & 6 & \textbf{19} & 1.551 & 32 & 56.6 \\
Qwen3.8 Max & 6 & \textbf{20} & 1.545 & 24 & 58.1 \\
GPT-6 Astra (medium) & 6 & 22 & 1.539 & \textbf{19} & 59.2 \\
Grok 4.6 (xhigh) & 6 & 24 & 1.529 & \textbf{16} & 60.0 \\
GPT-5.6 Sol (xhigh) & 9 & 27 & 1.520 & \textbf{20} & 59.0 \\
Grok 4.6 (medium) & 6 & 30 & 1.493 & \textbf{20} & 59.0 \\
\bottomrule
\end{tabular*}
\end{table}

\section{External validation and reporting analysis}
\label{sec:app-external}

We test retention, semantic weighting, and reporting on one previously unused
BIG-bench matrix. The original development had no reserved validation corpus.
Earlier \corpusname{} results informed the questions and candidate adoption rule.
We set the design on 7 November 2026 before comparative fitting or semantic-outcome
analysis on BIG-bench. The subanalyses share data and configurations.
Appendix Details, \S\,I gives protocols, figures, and secondary tables.

\subsection{Source, score policy and paired evaluation}

We select zero-shot scores with documented temperature-zero decoding from
BIG-bench revision \texttt{092b196}. This policy yields 12 BIG-G dense, ten
BIG-G sparse, and eight GPT configurations. For each leaf, we require
examples, a preferred metric of \texttt{exact\_str\_match} or
\texttt{multiple\_choice\_grade}, and scores for at least 24 configurations.
We use raw fractions for both metrics. For each parent, we average a fixed
set of eligible leaves and record a score only for configurations with
all required leaf scores. Before fitting, we exclude
\texttt{goal\_step\_wikihow} because it contains conflicting records for one
configuration. The resulting matrix contains \bobExternalLeaves{} leaves
and \bobExternalTasks{} parents, with \bobExternalCoverage{}\% of parent
cells observed.

We keep each parent and all its leaves in the same source block and assign
the blocks to five folds for withholding. No cross-parent input-reuse link
meets the threshold in \Cref{sec:app-external-semantics}. Within each fold,
we exclude constant columns from training and require at least 20 observed
scores and nonzero variation in held-out targets. These criteria yield
\bobExternalTargets{} comparable targets on the same \bobExternalModels{}
configurations.

For semantic weighting, we embed official task names and descriptions with
Qwen3-Embedding-8B and measure angular distances. Within each paired
comparison, we use a common bandwidth of one quarter of the unperturbed
training median distance. After fitting for six rounds, we compute
correlations across individual configurations and average first over suite
seeds, then over targets within source blocks, and finally over equally
weighted blocks. For retention intervals, we resample source blocks
conditional on the observed configurations. We apply Holm correction to
the two primary retention tests and the full-pool BoB-minus-equal contrast.

The external comparison uses a separate score matrix with shared task
lineage. \corpusname{} includes BBH and BBEH, and BIG-bench includes source
tasks for BBH. For the input-reuse analysis, we compare the public task
definitions. The public files document evaluation metadata, with the
correspondence between those definitions and each model's evaluated items
unverified.

\subsection{Retention across suite policies and composition}

We sample across ordered keyword domains with five fixed seeds to construct
nested balanced suites, starting with ten tasks. To compare suite
compositions, we include uniform random and single-domain suites as well as
the eligible BIG-bench Lite v1 intersection. For the Lite comparison, we
match the uniform random and balanced suites to the number of Lite tasks
retained for training in each fold.

BoB's held-out Spearman rises by $0.0354$ [$0.0221$, $0.0489$] as the
balanced pool grows from ten to 25 tasks, and by a further $0.0283$
[$0.0145$, $0.0419$] from 25 to 50. With suite size fixed at ten tasks, the
balanced suite's held-out Spearman exceeds that of the uniform random suite
by $0.0420$ [$0.0161$, $0.0697$]. At that same size, the balanced suite's
differences from the single-domain suites range from
$-0.0119$ [$-0.0401$, $0.0169$] for multilingual to
$+0.0215$ [$+0.0003$, $+0.0427$] for reasoning.

\begin{table}[H]\centering\small
\caption{Full-pool minus subset Spearman under fixed suite policies. BoB
intervals are exploratory 95\% source-block bootstrap intervals. Lite-matched
sizes vary by held-out fold.}
\label{tab:external-policies}
\begin{tabular}{@{}llrr@{}}
\toprule
Suite policy & Tasks & BoB retention [95\% CI] & Equal retention \\
\midrule
Domain balanced & 10 & +0.0825 [+0.0581, +0.1066] & +0.0787 \\
Domain balanced & 25 & +0.0471 [+0.0269, +0.0674] & +0.0422 \\
Domain balanced & 50 & +0.0188 [+0.0088, +0.0290] & +0.0158 \\
Uniform random & 10 & +0.1244 [+0.0843, +0.1655] & +0.1208 \\
Uniform random & 25 & +0.0660 [+0.0378, +0.0948] & +0.0614 \\
Uniform random & 50 & +0.0165 [+0.0061, +0.0270] & +0.0099 \\
BIG-bench Lite & 15--19 & +0.0485 [+0.0300, +0.0662] & +0.0444 \\
Random, Lite-matched & 15--19 & +0.0876 [+0.0548, +0.1210] & +0.0835 \\
Balanced, Lite-matched & 15--19 & +0.0603 [+0.0392, +0.0812] & +0.0561 \\
\bottomrule
\end{tabular}

\end{table}

To assess retention within architecture families, we recompute correlations
for each of the three families using abilities from the shared fits. BoB's
retention estimates are positive in all three families, ranging from
$+0.0461$ in GPT to $+0.1032$ in BIG-G sparse. All six method-by-family
intervals exclude zero. Appendix Details, \S\,I provides the nested curves
and the within-family results (original Figure 12 and Table 53).

\subsection{Item reuse and score-side association}
\label{sec:app-external-semantics}

To measure reuse in task definitions, we hash normalized string inputs of at
least 20 characters before inspecting embeddings. A supported pair requires
at least ten unique retained inputs per leaf. We label exact-input reuse at
five or more shared inputs and at least 0.10 overlap relative to the smaller
set. Zero intersection defines the negative label. Within parents, 178 pairs
from seven parents are positive, 9,855 have zero intersection, 768 are
intermediate, and 25,736 lack support. Average precision from description
distance is \bobExternalReuseAP{} against prevalence \bobExternalReusePrevalence{}.
The cross-parent data contain zero supported positives. Testing question reuse
in \corpusname{} would require item-level records beyond its aggregate scores
and provenance.

For the score-side analysis, we fit a quadratic trend in a pair-excluded
ability proxy on other model architecture families and remove it. Among 8,099
eligible parent pairs, the association of semantic similarity with cross-fitted
residual correlation is \bobExternalResidualAssociation{}
[\bobExternalResidualLo{}, \bobExternalResidualHi{}], with raw-score
association 0.0199 [$-0.0584$, 0.1007]. The node-permutation test gives
$p=0.502$ before adjustment and 1.000 with Holm adjustment for two planned
tests. The observed association is compatible with zero.
Benchmarks that cover related capabilities can share little weight when
their descriptions are distant in the embedding space. Composite-indicator
guidance treats indicator content as evidence about double counting alongside
observed correlations \citep[p.~33]{nardo2008handbook}. The weights in
\Cref{sec:weighting} use descriptions and reporting indicators.

We add a second disjoint supported leaf in eight parents to test related
entries with new evidence. The resulting incremental Spearman
difference between BoB and equal equating is $-0.0015$ [$-0.0056$, $+0.0014$].
For the matched copy, mean displacement is 0.017 under BoB and 0.242 under
equal equating. Appendix Details, \S\,I gives the
paired effects, resampling, and plots (original Figures 13--14).

\paragraph{Retrieval of related task designs.}
Average precision favors description embeddings for documented BBH--BBEH
task links and character TF--IDF for shared inputs within BIG-bench
(\cref{tab:appendix-lineage}). For each of 23 BBEH tasks, we
rank all 23 BBH candidates using the published descriptions after masking task
titles and explicit references to the predecessor \citep{suzgun2023bigbenchhard,kazemi2025bbeh}.
Each query has one documented predecessor. We pool the 529 pairs for average
precision (AP). In the BIG-bench check, both methods score the same 10,033
labeled within-parent pairs. In both checks, we use character $n$-grams of
length three to five for the TF--IDF baseline. The BBH--BBEH links identify which tasks were
designed as successors. Such pairs can contribute distinct evidence at
different levels of difficulty. Other kinds of overlap are unlabelled.

\begin{table}[H]
\centering\small
\caption{Average precision for retrieving shared inputs and documented task
predecessors. Parentheses give the Qwen3-Embedding model size.}
\label{tab:appendix-lineage}
\begin{tabular}{lrr}
\toprule
Retrieval target & Character TF--IDF & Description embeddings \\
\midrule
BIG-bench shared inputs & 0.276 & 0.146 (8B) \\
BBH--BBEH task predecessors & 0.415 & 0.537 (0.6B), 0.485 (4B) \\
\bottomrule
\end{tabular}

\end{table}

\subsection{Column weighting under partial reporting}
\label{sec:app-external-reporting}

Rank displacement under the original \corpusname{} column rule exceeds that
under equal equating in 11 of 24 quarter-coverage copy scenarios. These include
all three masks each for AIME 2025 and GPQA Diamond, two each for SWE-bench
Verified and BFCL v3, and one for MMMU validation. Over eight
sources, column density minus equal displacement averages $+0.1758$ for
114 models with 3--9 reports, $+0.0468$ for 213 models with 10--29 reports,
and $-0.1049$ for 27 models with at least 30. The reporting counts precede
copy addition. These are descriptive strata.

In the external grid, we add one or four copies of a source column at 25\%,
50\%, 75\%, or 100\% of observed rows, with shared, independent, or complementary
masks. We use three seeds at partial coverage and all observed rows at
complete coverage. We assign six sources each to development and validation,
hold out 25 variable targets, and conduct inference at the source level.
Each partition contains 336 scenarios.
Appendix Details, \S\,I plots the support-stratified effects and the
coverage curves (original Figures 15--16).

\subsection{An observable rule for controlled re-listings}
\label{sec:app-external-rule}

We fix the candidate rule before external validation. It requires a nonempty
set of edges with kernel similarity
at least 0.5, both directed conditional reporting overlaps at least 0.50
for every such edge, and a tenth-percentile observed mass of at least ten.
Observed mass is $M_m=\sum_i O_{mi}u_i$, with $O_{mi}$ the observation
indicator and $u_i$ the unnormalized coefficient in \Cref{eq:mass}.
The rule uses descriptions and reporting indicators.

The rule selects \bobExternalRuleSelected{} of \bobExternalRuleTotal{}
validation scenarios. Displacement averages 0.0559 under density equating
and 0.2210 under equal equating, a difference of
\bobExternalRuleRankDelta{} [\bobExternalRuleRankLo{}, \bobExternalRuleRankHi{}].
Five of six source means favor density (one-sided sign-flip $p=0.03125$),
and \bobExternalRuleHarm{} selected scenarios favor equal equating. The
rejected regime's mean difference is $-0.0500$, with 40 of 144 scenarios
favoring equal weighting.

The selected-regime predictive difference is \bobExternalRuleUtilityDelta{}
[\bobExternalRuleUtilityLo{}, \bobExternalRuleUtilityHi{}]. The one-sided
95\% lower bound of \bobExternalRuleUtilityLower{} exceeds the declared
$-0.01$ Spearman noninferiority margin. Holm-adjusted $p$ values for
stability and noninferiority are 0.03125 and 0.0022. These tests use six
intervention sources and 25 target blocks. The sign-flip test assumes the
source-effect signs are exchangeable under the null.

Observed mass at the tenth percentile is 90.36--91.83 in this dense grid,
so the mass criterion never binds. The unchanged rule fails both conditions
on the unperturbed \corpusname{} panel, with minimum conditional overlap
zero and tenth-percentile mass 4.02. The unperturbed BIG-bench panel passes
both. Leaving out any source preserves the negative mean effect, while five
of the six one-sided $p$ values become 0.0625. These sensitivity checks
locate the evidence in the six-source panel and the tested overlap range
(Appendix Details, \S\,I, original Figure 17).

\subsection{Outcome-dependent reporting sensitivity}
\label{sec:app-external-selective}

We sample training cells without replacement with weights
$\exp(\beta z_{mi})$, where $z_{mi}$ is the known score standardized within
its column. Positive tilt favors high scores, negative tilt favors low
scores, and zero tilt is random. At each coverage level, we hold the raw
budget fixed across tilts and use the same 25 targets and 30 models in all
90 fits. Filtering constant columns can change fitted support.

At quarter coverage, held-out Spearman for BoB falls from 0.1735 under
random masking to 0.1124 with $\beta=2$ and 0.0739 with $\beta=-2$.
Equal equating has similar losses. BoB's displacement is 3.689 ranks under
random masking and ranges from 5.000 to 6.578 under the two extreme tilts.
These effects measure sensitivity to the specified mechanisms on known cells.
The actual unreported outcomes in \corpusname{} remain unidentified.

\begin{table}[H]\centering\footnotesize
\caption{Outcome-dependent reporting stress test, averaged over three masks
per setting. Displacement compares each masked fit with the same estimator's
unmasked ranking. Tilt preserves each column's raw observation budget.
\bobname{} denotes the column form.}
\label{tab:external-selective}
\begin{tabular}{@{}rrrrrr@{}}
\toprule
Nominal & $\beta$ & BoB $\rho$ & Equal $\rho$ & BoB $\Delta\rho$ vs. random & BoB displacement \\
\midrule
25\% & -2 & 0.0739 & 0.0737 & -0.0996 & 6.578 \\
25\% & -1 & 0.1089 & 0.1131 & -0.0646 & 5.511 \\
25\% & +0 & 0.1735 & 0.1730 & +0.0000 & 3.689 \\
25\% & +1 & 0.1190 & 0.1208 & -0.0545 & 5.244 \\
25\% & +2 & 0.1124 & 0.1172 & -0.0610 & 5.000 \\
50\% & -2 & 0.1343 & 0.1333 & -0.0778 & 5.089 \\
50\% & -1 & 0.1796 & 0.1819 & -0.0324 & 3.467 \\
50\% & +0 & 0.2121 & 0.2055 & +0.0000 & 2.178 \\
50\% & +1 & 0.1432 & 0.1389 & -0.0689 & 3.667 \\
50\% & +2 & 0.1331 & 0.1249 & -0.0789 & 3.889 \\
75\% & -2 & 0.1881 & 0.1888 & -0.0177 & 2.867 \\
75\% & -1 & 0.2014 & 0.2021 & -0.0044 & 2.311 \\
75\% & +0 & 0.2058 & 0.2014 & +0.0000 & 1.311 \\
75\% & +1 & 0.1741 & 0.1795 & -0.0317 & 2.667 \\
75\% & +2 & 0.1480 & 0.1438 & -0.0578 & 3.378 \\
\bottomrule
\end{tabular}

\end{table}

Appendix Details, \S\,I plots the paired effects (original Figure 18).
The study archive records the masks and counts.

\subsection{Transfer of WildScores reporting masks}
\label{sec:app-real-masks}

We map 143 BIG-bench parent descriptions to distinct \corpusname{} columns
with score-blind one-to-one maximum-cosine assignment (median 0.481, range
0.255--0.707). In each of 24 draws, we sample 30 WildScores reporting rows
without replacement and transfer their binary masks to BIG-bench. Paired
column permutation preserves row counts, and native missing cells remain absent.
Within five source-block folds, the three-observation filter retains 23--30
models and 25--69 columns under aligned masks. We match dense and masked fits
on retained rows and columns and set bandwidth from unfiltered training geometry.
We correlate fitted global ability with held-out scores on 129 variable targets
shared by all methods and draws. This comparison measures sensitivity to
co-reporting structure transferred across models and benchmarks. Outcome-dependent
bias in WildScores remains unidentified.

Mean held-out Spearman for BoB falls from \bobSparseDenseReference{} to
\bobSparseDensity{} under aligned masks. Under equal equating it is
\bobSparseEqual{}, with a density-minus-equal difference of \bobSparseGap{}
[\bobSparseGapLo{}, \bobSparseGapHi{}]. Permutation gives the same qualitative
result. The adoption rule in \Cref{sec:app-external-rule} fails for every
fitted mask under both mappings.

\begin{table}[H]\centering\footnotesize
\caption{Transfer of observed reporting masks. Dense and masked fits match
in retained models and training columns. Values average Spearman between
$\theta$ and held-out scores. \bobname{} denotes the column form. Intervals
resample draws and target blocks.}
\label{tab:real-masks}
\resizebox{\linewidth}{!}{\begin{tabular}{@{}lrrrrr@{}}
\toprule
Mapping & Dense BoB & Dense equal & Masked BoB & Masked equal & Difference [95\% interval] \\
\midrule
Aligned & 0.2298 & 0.2301 & 0.0843 & 0.0852 & $-0.0009$ [$-0.0041$, $+0.0022$] \\
Permuted & 0.2300 & 0.2306 & 0.1079 & 0.1099 & $-0.0020$ [$-0.0058$, $+0.0022$] \\
\bottomrule
\end{tabular}
}
\end{table}

\subsection{Density over reported benchmarks}
\label{sec:app-mask-aware}

In the reporting-aware form of \Cref{sec:weighting}, we compute model-specific
coefficients $u_{mi}=[\sum_j O_{mj}\exp(-d_{ij}^2/(2h^2))]^{-1}$ for observed
$i$. We change the ability update and hold the six rounds, percentile
initialization, response curves, robust residual weights, and bandwidth
fixed across rules. A copy enters a model's density only if the model reports it.
For $t$ observed copies with other neighbors contributing
kernel mass $C_m$, the group has raw mass $t/(t+C_m)$. An isolated group
therefore has mass one. Other neighbors and score-side refitting can affect
rankings. For the fourth rule, we use the class-uniform graph construction of
\citet{berriaud2026consensus}, with the exact uniform-radius integral and
locality radius $2h$. This specifies a member of their weighting family
separate from the Euclidean-ball construction \citep{berriaud2026clone}.

\begin{table}[H]\centering\small
\caption{Comparison of column density and reporting-aware (model) density.
The copy rows average all scenarios at each coverage. The affine row first
averages seeds within sources across the 17 cases where all rules converge.
Predictive rows match models, targets, and training pools. BW is the
class-uniform column rule.}
\label{tab:mask-aware}
\resizebox{\linewidth}{!}{\begin{tabular}{@{}lrrrr@{}}
\toprule
Condition & Equal & Column density & Model density & BW columns \\
\midrule
\multicolumn{5}{l}{Nonlinear copy displacement (lower is better)} \\
Complete & 1.9484 & 0.6250 & 0.2698 & 0.1405 \\
25\% & 0.6840 & 0.7607 & 0.1508 & 0.6548 \\
50\% & 1.2032 & 0.7056 & 0.1969 & 0.4952 \\
75\% & 1.6095 & 0.6569 & 0.2427 & 0.3274 \\
Reporting pattern & 1.0890 & 0.7267 & 0.1716 & 0.5275 \\
\midrule
\multicolumn{5}{l}{Affine displacement (common converged cases, lower is better)} \\
25\% & 0.6414 & 0.6056 & 0.2408 & 0.5581 \\
\midrule
\multicolumn{5}{l}{Global held-out Spearman (higher is better)} \\
Aligned real masks & 0.0852 & 0.0843 & 0.0852 & 0.0844 \\
Permuted real masks & 0.1099 & 0.1079 & 0.1089 & 0.1095 \\
WildScores strict, full & 0.8402 & 0.8386 & 0.8394 & 0.8393 \\
WildScores broad, full & 0.7994 & 0.8016 & 0.7998 & 0.7983 \\
\bottomrule
\end{tabular}
}
\end{table}

Rank displacement under model density is smaller than under equal equating
in all 88 nonlinear copy scenarios. At 25\%, the paired effect is \bobMaskCopyGap{}
[\bobMaskCopyLo{}, \bobMaskCopyHi{}] with Holm-adjusted $p=\bobMaskCopyHolm{}$.
At 25\%, mean displacement is lower under BW, and seven cases have larger
displacement.
The affine comparison reproduces the mean benefit (\bobMaskAffineGap{}
[\bobMaskAffineLo{}, \bobMaskAffineHi{}], Holm $p=\bobMaskAffineHolm{}$),
with two individual reversals at 25\% coverage. Across the four affine
rules, we use unweighted column regression and a common fixed-point stopping
rule distinct from the original affine comparison. We fit from three starts, with up to
20,000 iterations, and thresholds of $10^{-7}$ for ability changes and
relative reconstruction changes over three consecutive iterations. Failures
among 89 fits per rule number four for equal, three each for column and
model density, and 21 for BW. We pair cases where all rules converge.

The aligned real-mask difference from equal equating is
\bobMaskPredictionGap{} [\bobMaskPredictionLo{}, \bobMaskPredictionHi{}].
In the strict \corpusname{} full pool the difference is
\bobMaskRetentionGap{} [\bobMaskRetentionLo{}, \bobMaskRetentionHi{}], meeting
the specified 0.01-Spearman noninferiority margin. Full-minus-curated gains
remain positive under model density in both retention panels. We average copy
seeds before source resampling, cross draws with source blocks for mask
resampling, and resample source families for retention. We apply Holm
correction jointly to the three directional efficacy tests. The tests support stability
under repeated listings with strict-panel agreement within the declared
margin. Across the four rules in \Cref{tab:mask-aware}, weighting
changes agreement less than expanding score coverage in either retention
panel (\Cref{tab:insight-comparison}). This is consistent with the classical
analysis of correlated composites \citep{wilks1938weighting}.
The reporting-aware study in the study archive records the affine
selection and failure protocol and reconstruction checks.

\subsection{Natural re-listings}
\label{sec:app-natural}

\paragraph{Shared questions in a published composite.}
The default OpenVLM Leaderboard averages eight benchmark scores on a 0--100
scale, including MathVista testmini and MMStar \citep[Sec.~3]{duan2024vlmevalkit}.
MMStar selects questions that require visual input and takes 345 of its 1,500
items from MathVista \citep[Fig.~3(a)]{chen2024mmstar}. MathVista testmini
item 404 and MMStar item 399 present the same timetable, question, and four
choices \citep{lu2023mathvista,chen2024mmstar}.\footnote{The leaderboard defaults
are defined at \url{https://huggingface.co/spaces/opencompass/open_vlm_leaderboard/blob/main/meta_data.py}.
The data releases are \texttt{AI4Math/MathVista} at \texttt{2b6ad69} and
\texttt{Lin-Chen/MMStar} at \texttt{bc98d66}.} The question contributes to both
scores in the composite. In the original benchmark-score model behind the
Epoch Capabilities Index, splitting the same questions into two entries of
similar difficulty can roughly double their combined influence
\citep[Sec.~4.4]{ho2025rosetta}.

\paragraph{Third-party measurements.}
Artificial Analysis measures thirteen \corpusname{} benchmarks with its own
harness. We match added scores to any eligible developer counterpart a model
reports, with possible differences in evaluation settings. Their weight share
under equal equating rises from \bobNatShareEqualBase{}\% to
\bobNatShareEqualRemeasure{}\%. Model density moves the share from
\bobNatShareModelBase{}\% to \bobNatShareModelShared{}\% when both sources
share a description and to \bobNatShareModelSeparate{}\% with separately
written descriptions. Corresponding rank displacement is \bobNatDispEqual{}
under equal equating and \bobNatDispModelShared{} or \bobNatDispModelSeparate{}
under model density. Lifting the counterpart requirement adds mostly first
measurements and raises the share to \bobNatShareFullLow{}--\bobNatShareFullHigh{}\%
across rules. On that larger
pool, model density's held-out difference from equal equating is
\bobNatStrictGapShared{} [\bobNatStrictGapSharedLo{}, \bobNatStrictGapSharedHi{}]
in the strict panel and \bobNatBroadGapShared{}
[\bobNatBroadGapSharedLo{}, \bobNatBroadGapSharedHi{}] in the broad panel with
a shared description. The separate-description effects are
\bobNatStrictGapSeparate{} [\bobNatStrictGapSeparateLo{}, \bobNatStrictGapSeparateHi{}]
and \bobNatBroadGapSeparate{} [\bobNatBroadGapSeparateLo{}, \bobNatBroadGapSeparateHi{}].
The targets are developer-reported scores from corpora used in development.
\Cref{sec:app-mask-aware} gives the prediction interval for the unaugmented
strict panel.

\begin{table}[H]\centering\small
\caption{Third-party measurements of thirteen \corpusname{} benchmarks.
Share is their percentage of total model weight. Displacement compares with
the fit without the new columns and averages absolute rank changes over 354
models. Held-out values are family-mean Spearman. Model density denotes the
reporting-aware form. BW is the class-uniform column rule.}
\label{tab:natural-remeasure}
\resizebox{\linewidth}{!}{\begin{tabular}{@{}lrrrrrrr@{}}
\toprule
 & \multicolumn{3}{c}{Share (\%)} & \multicolumn{2}{c}{Displacement} & \multicolumn{2}{c}{Strict Spearman} \\
\cmidrule(lr){2-4}\cmidrule(lr){5-6}\cmidrule(l){7-8}
Rule & Before & Separate & Shared & Separate & Shared & Separate & Shared \\
\midrule
\multicolumn{8}{l}{Re-measured cells only} \\
Equal & 33.8 & 44.1 & 44.1 & 5.61 & 5.61 & 0.8358 & 0.8358 \\
Column density & 30.9 & 37.3 & 34.0 & 4.73 & 4.08 & 0.8317 & 0.8380 \\
Model density & 33.7 & 39.4 & 35.9 & 4.95 & 4.20 & 0.8343 & 0.8417 \\
BW columns & 33.4 & 42.3 & 34.2 & 5.47 & 3.83 & 0.8323 & 0.8399 \\
\midrule
\multicolumn{8}{l}{Full third-party columns} \\
Equal & 33.8 & 59.9 & 59.9 & 11.29 & 11.29 & 0.8363 & 0.8363 \\
Column density & 30.9 & 53.1 & 48.5 & 10.60 & 9.97 & 0.8359 & 0.8367 \\
Model density & 33.7 & 57.9 & 56.6 & 11.49 & 11.76 & 0.8325 & 0.8386 \\
BW columns & 33.4 & 58.1 & 47.5 & 11.31 & 9.52 & 0.8347 & 0.8371 \\
\bottomrule
\end{tabular}
}
\end{table}

\paragraph{Sharing weight across confirmed re-measurements.}
When benchmark identities are known, we compute reporting-aware density over
the distinct identities a model reports. For observed identity $g$, we divide
its nominal weight $u_{mg}$ equally among the $n_{mg}$ available source scores.
All scores enter the fit, with a separate response curve per source column.
When the observed identities and geometry stay fixed, the nominal weight and
share are invariant to an added measurement. The ranking can change after
refitting.

We apply a stricter match than in \cref{tab:natural-remeasure}, requiring an
observed score in the designated developer counterpart for that model.
The match yields 824 additional Artificial Analysis scores over thirteen
benchmarks. We fit all three rules on the same augmented pool
(\cref{tab:appendix-identity}). Held-out Spearman differences between the
proposed rule and reporting-aware \bobname{} with
shared descriptions are $+0.00012$
[$-0.00106$, $+0.00119$] in the strict panel and $-0.00016$
[$-0.00063$, $+0.00031$] in the broad panel (95\% family-bootstrap intervals).

\begin{table}[H]\centering\small
\caption{Average nominal weight share of the thirteen benchmarks in the
stricter same-model, same-benchmark comparison. The reporting-aware rule uses
shared descriptions for source scores of the same benchmark.}
\label{tab:appendix-identity}
\begin{tabular}{lrrr}
\toprule
Weighting rule & Share before & Share after & Change (pp) \\
\midrule
Equal per column & 33.80\% & 41.68\% & +7.88 \\
Reporting-aware \bobname{} & 33.66\% & 34.01\% & +0.35 \\
Weight split by benchmark identity & 33.66\% & 33.66\% & 0.00 \\
\bottomrule
\end{tabular}

\end{table}

\paragraph{Benchmark families.}
\bobNatFamilies{} families with multiple listings account for
\bobNatFamilyColumns{} of 148 columns. For each family, we keep one listing
and refit at a fixed bandwidth. Mean displacement under model density is
smaller than under equal equating in \bobNatFamiliesModelSmaller{}
families, more in \bobNatFamiliesModelLarger{}, and ties in
\bobNatFamiliesModelTies{}. Keeping one RULER length at a time, models that
originally report at least two lengths move \bobNatRulerEqual{},
\bobNatRulerColumn{}, and \bobNatRulerModel{} ranks on average under equal,
column-density, and reporting-aware weights. For AIME the three rules give
\bobNatAimeEqual{}, \bobNatAimeColumn{}, and
\bobNatAimeModel{}. Displacements for families with distinct task sets,
such as tau-bench and SWE-bench variants, are similar across rules. The
displacements reflect both the change in listing count and the observations
lost with each listing. Shares change little in the largest broad
categories. Appendix Details, \S\,I gives all family results, RULER subsets,
and reporter restrictions (original Tables 58--59).

\section{Close model comparisons and target-score estimation}
\label{sec:app-close-comparisons}

Pooling scores can reduce dependence on any one measurement. Near the ceiling,
close comparisons still need benchmarks that distinguish leading models.

\subsection{Close model pairs on Artificial Analysis}
\label{sec:app-frontier-pairs}

Agreement with held-out benchmark orders improves as we add scores for closely
ranked recent checkpoints under both equating rules (\cref{tab:appendix-frontier}).
We use the November 18 Artificial Analysis v5.0.1 snapshot of 57 configurations
with complete scores on nine benchmarks. Keeping the highest reported inference
effort per checkpoint yields 30 calibration checkpoints, including 19 released
on or after October 1, 2026.

For each held-out benchmark, we select pairs within three positions of each
other in the upper half of the recent checkpoints, ranked by their mean
percentile on the other eight benchmarks. We evaluate all four fits on the
same 209 pair--benchmark comparisons after excluding held-out ties. We curate
the pool by taking the highest-priority available benchmark in each of four
published AA categories \citep{artificialanalysis2026methodology}. We average concordance
over the nine targets, with half credit for predicted ties. The median held-out
gap is 6.5 percentage points. Both weighting rules improve on six targets, tie
on one and decline on two when the pool expands. These estimates describe
agreement with reported scores as the number and composition of predictors vary.

\begin{table}[H]
\centering\small
\caption{Mean agreement with held-out benchmark orders for closely ranked
recent checkpoints. Each column uses the same model pairs and targets.}
\label{tab:appendix-frontier}
\begin{tabular}{lrr}
\toprule
Equating weights & Four benchmarks & Eight benchmarks \\
\midrule
Equal & 68.6\% & 70.6\% \\
\bobname{} & 68.6\% & 71.1\% \\
\bottomrule
\end{tabular}

\end{table}

\paragraph{Repeatability near the ceiling.}
Across six splits of saved MathArena responses, the second half preserved
24 of 32 model orders from the first half and reversed eight. We average HMMT,
AIME and BRUMO scores equally within each half of the four response positions
for 44 configurations. In the first half, both models in a selected pair scored
at least 171 of 180 (95\%) and differed by one correct response (0.56 percentage
points).
The splits span 20 distinct pairs and reuse responses. The reversals show
measurement uncertainty for sub-point gaps on these fixed questions.

\subsection{Estimating HMMT accuracy from nearby and distant scores}
\label{sec:app-hmmt}

With five scores in the pool, HMMT estimation error is lower under
relevance and density weighting than under equal weighting
(\cref{tab:appendix-hmmt}). We use MathArena's February 2025 HMMT, AIME 2025
and BRUMO 2025, with archived IFBench and AA-LCR scores from Artificial Analysis.
AIME and BRUMO lie 0.50 and 0.52 radians from the HMMT description. IFBench and
AA-LCR lie 1.03 and 1.06 radians away. We match 21 configurations in six model groups by
published checkpoint and reasoning mode. Historical inference settings across
providers are partially observed.

\paragraph{Paired comparison.}
In each of four rotations, we use one saved HMMT response per question.
The other three responses on each of 30 questions give a 90-grade reference
mean. We share monotone maps across rules to calibrate auxiliary scores to
the same input position on other model groups. In this comparison, \bobname{}
averages HMMT and the calibrated scores using Gaussian relevance to HMMT
times inverse description density, normalized to sum
to one. The respective bandwidths are 0.45 and 0.25 times the median angular
distance in \corpusname{}. For the matched control, we hold the HMMT weight
fixed and use inverse density alone for the remaining scores.

\begin{table}[H]
\centering\small
\caption{Weight shares and held-out mean squared error for HMMT estimation.
Variance is the mean across configurations of the sample variance over the four
input positions. Error and variance are in squared percentage points.}
\label{tab:appendix-hmmt}
\begin{tabular}{lrrrrr}
\toprule
Weighting rule & HMMT & AIME + BRUMO & IFBench + AA-LCR & MSE & Variance \\
\midrule
Equal & 20.0\% & 40.0\% & 40.0\% & 78.20 & 6.83 \\
Fixed HMMT, inverse density & 47.3\% & 23.9\% & 28.8\% & 50.65 & 9.99 \\
\bobname{} & 47.3\% & 46.7\% & 6.0\% & 32.93 & 12.09 \\
\bottomrule
\end{tabular}

\end{table}

Across HMMT input positions, variance is lower under equal weighting. Mean
squared error is lower under \bobname{} by 45.27 squared percentage points,
with a paired 95\% interval of $[1.51,141.24]$. At the same HMMT weight of
47.3\%, the estimated reduction from adding relevance to inverse density is
17.72, with an interval of $[-4.02,70.54]$. We resample model groups
conditional on questions, recorded responses and calibration fits. In the
complementary MATH-500 comparison of published scores, the absolute error
estimate rises when we add related scores
(\emph{Appendix Details}, target-score comparisons).

\end{document}